\documentclass[preprint,12pt]{elsarticle}

\usepackage{amssymb}
\usepackage{amsmath}

\usepackage{subcaption}
\usepackage{hyperref}
\usepackage{url}
\usepackage{wrapfig}
\usepackage{layouts}
\usepackage{float}
\usepackage{booktabs}
\usepackage{placeins}
\usepackage{cleveref}
\usepackage{xcolor}

\journal{Neurocomputing}

\begin{document}

\begin{frontmatter}



\title{A Practical Guide to Streaming Continual Learning}


\author[unipi]{Andrea Cossu\corref{equal}}\ead{andrea.cossu@unipi.it}
\author[polimi]{Federico Giannini\corref{equal}}\ead{federico.giannini@polimi.it}
\author[polimi]{Giacomo Ziffer}
\author[motus]{Alessio Bernardo}
\author[haw]{Alexander Gepperth}
\author[polimi]{Emanuele Della Valle}
\author[bie]{Barbara Hammer}
\author[unipi]{Davide Bacciu}

\cortext[equal]{Equal contribution}

\affiliation[unipi]{organization={Computer Science Department, University of Pisa},
            addressline={Largo Bruno Pontecorvo, 3}, 
            city={Pisa},
            postcode={56127}, 
            country={Italy}}
            
\affiliation[polimi]{organization={DEIB, Politecnico di Milano},
            addressline={Via Giuseppe Ponzio, 34}, 
            city={Milano},
            postcode={20133}, 
            country={Italy}}
            
\affiliation[haw]{organization={University of Applied Sciences Fulda}, country={Germany}}
\affiliation[bie]{organization={Bielefeld University}, country={Germany}}
\affiliation[motus]{organization={Motus ml}, 
            city={Milano}, 
            country={Italy}}

\begin{abstract}
Continual Learning (CL) and Streaming Machine Learning (SML) study the ability of agents to learn from a stream of non-stationary data. Despite sharing some similarities, they address different and complementary challenges. While SML focuses on rapid adaptation after changes (concept drifts), CL aims to retain past knowledge when learning new tasks. After a brief introduction to CL and SML, we discuss Streaming Continual Learning (SCL), an emerging paradigm providing a unifying solution to real-world problems, which may require both SML and CL abilities. We claim that SCL can i) connect the CL and SML communities, motivating their work towards the same goal, and ii) foster the design of hybrid approaches that can quickly adapt to new information (as in SML) without forgetting previous knowledge (as in CL). We conclude the paper with a motivating example and a set of experiments, highlighting the need for SCL by showing how CL and SML alone struggle in achieving rapid adaptation and knowledge retention. 
\end{abstract}


\begin{highlights}
\item We provide a short background on Streaming Machine Learning (SML) and Continual Learning (CL), highlighting their similarities and differences.
\item We define the Streaming Continual Learning (SCL) scenario, which offers the opportunity to merge ideas from CL and SML into a unified paradigm.
\item We conduct an empirical evaluation showcasing the need for Streaming Continual Learning (SCL).

This is the accepted manuscript of the article:\\
Cossu, A., Giannini, F., Ziffer, G., Bernardo, A., Gepperth, A., Della Valle, E., Hammer, B., Bacciu, D. (2026). \textbf{A practical guide to streaming continual learning}. Neurocomputing, 674, 132951.

The final published version is available at:\\ \url{https://doi.org/10.1016/j.neucom.2026.132951}\\
© 2026 Elsevier.

\end{highlights}


\end{frontmatter}



\section{Introduction}
Over the last decade, the exponential growth of interconnected digital systems, driven by platforms such as social media and the Internet of Things (IoT), has resulted in massive, continuous flows of information. This results in the production of massive amounts of data, often generated by unbounded flows called data streams \cite{book_bifet}. Data arrives sequentially and continuously over time, making it impractical to store or revisit them in full.

This streaming scenario challenges traditional machine learning paradigms, which often assume stationary data and bounded datasets. One of the most critical obstacles is concept drift~\cite{cit:cd_tsymbal2004}, a phenomenon where the statistical properties of the data evolve due to underlying changes in the generative process (\emph{concept}). Unlike sporadic anomalies, concept drift reflects structural, often unforeseeable changes that can render previously learned models unsuitable. 

Due to concept drifts, the solution cannot train a model offline on historical data, since the trained model may soon become obsolete after concept drifts. Conversely, one must continuously learn from the data stream, monitoring and detecting changes, and adapting to them as rapidly as possible. Moreover, when learning new concepts, the model may forget what it learned regarding the previous ones. This issue is particularly critical when the new concept just expands the learning problem in a new subdomain, without contradicting what has been observed during the previous concepts. In this case, the model should be able to encapsulate the knowledge associated with the new concept while retaining the previously acquired one. This idea is associated with the well-known \textbf{stability-plasticity dilemma}~\cite{carpenter1986, cit:stability_plasticity,cit:stability_plasticity2}, according to which the model should achieve a trade-off between the ability to learn new knowledge (plasticity) and the ability to remember the previous (stability). Too much plasticity leads to forgetting, too much stability prevents learning new knowledge.

However, forgetting may be desirable in specific situations. When the new concept contradicts a previously learned one, then the model should not preserve both. In this situation, avoiding forgetting takes on a more nuanced meaning. It does not require remembering the entire history of previous concepts, but rather preserving an up-to-date representation of the current knowledge while adapting dynamically to changes. Accessing prior knowledge remains a useful skill, as new changes in the data stream might require quick remembering of previously seen concepts. \\
\begin{wrapfigure}{l}{0.4\textwidth}
\centering
\includegraphics[width=0.38\textwidth]{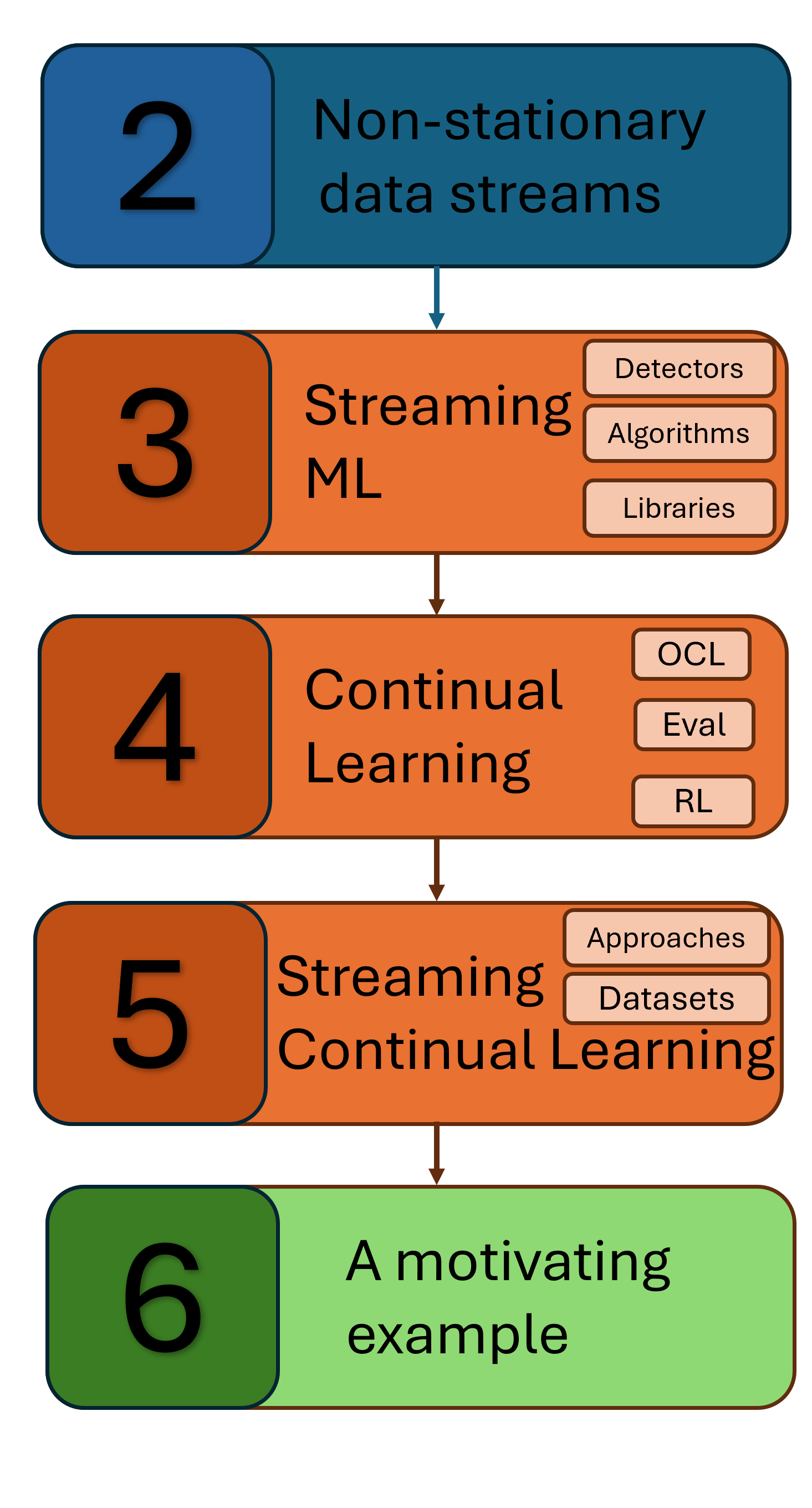}
\caption{Overview of the structure of this manuscript. We start with an introduction to non-stationary environments, followed by a short background on streaming and continual learning, before discussing the streaming continual learning paradigm together with an empirical example motivating why it is needed.}
\label{fig:overview}
\end{wrapfigure}

Two main research areas deal with the mentioned challenges. \textbf{Streaming Machine Learning (SML)}~\cite{book_bifet} focuses on producing resource-efficient solutions able to monitor and detect changes, and quickly adapt to the associated new concepts. SML prioritizes rapid adaptation over retaining previous knowledge. Its primary objective is to perform well on the current concept, even at the cost of forgetting past information. There is no explicit concern for preserving previously acquired knowledge. Instead, the system is designed to adapt swiftly to evolving data. If a new concept later requires reusing past knowledge, the model will simply relearn it from scratch. Conversely, \textbf{Continual Learning (CL)}~\cite{cit:cl} specifically prioritizes the preservation of the previously acquired knowledge while learning new concepts over quick adaptation to new concepts. These solutions, however, are meant to work in settings where new concepts just expand the problem with additional subdomains, without contradicting what was observed during previous concepts. 

Starting with these premises, this work highlights a clear gap in the literature. SML and CL are both important and complementary areas of research. However, on their own, they are not sufficient to provide a complete solution to real-world problems that demand continuous learning, rapid adaptation, and the ability to avoid forgetting (we further elaborate on this point in Section \ref{sec:example}). \textbf{Streaming Continual Learning (SCL)} aims at unifying, without replacing, CL and SML. Recently, the ideas and challenges of SCL are gaining traction and fostering a positive discussion in the research community. SCL has been initially suggested in \cite{cit:scl_gunasekara}, followed up in \cite{cit:scl}, and the topic of a dedicated special session at the European Symposium on Artificial Neural Networks (ESANN) in 2025~\cite{cossu2025a}.\\
This paper builds upon the ideas proposed in the tutorial paper from the aforementioned special session~\cite{cossu2025a}. In this extended version, we provide a more comprehensive overview of the scenarios, the objectives, and the evaluation protocols commonly used nowadays in CL and SML. We also expand the discussion about available datasets that can be directly used or repurposed to match the SCL framework. Finally, we added a motivating example showing the need for SCL: popular CL and SML approaches struggle in achieving both the CL objective (knowledge consolidation) and the SML objective (quick adaptation). \\
The paper also introduces the concept of a unified approach and discusses how this framework could evolve in the future by following specific research directions. Particularly, Section~\ref{sec:non_stationary_ds} presents the streaming scenario and the associated problem of concept drift. Section~\ref{sec:sml} presents SML, while Section~\ref{sec:cl} describes CL, by also illustrating the applications of Reinforcement Learning (RL) in streaming scenarios involving concept drifts. Section~\ref{sec:scl} introduces the vision of SCL and its motivation, while Section~\ref{sec:example} illustrates an example supporting the motivations behind SCL. Finally, Section~\ref{sec:conclusion} discusses the conclusion and the future directions we envision for SCL. Figure \ref{fig:overview} provides an overview of the structure of the paper.

\section{Non-stationary data streams}\label{sec:non_stationary_ds}
The notion of a \emph{data stream} emerges in two distinct scenarios, both involving a data source that is never fully accessible all at once. The first scenario, which we call \textbf{natural streams}, involves data that is generated continuously and in real-time. A typical example is the IoT, where sensors continuously gather measurements. In such cases, the model processes the current data instance (possibly along with a small buffer of recent past instances) and generates predictions accordingly. The second scenario, which we refer to as \textbf{artificial streams}, arises due to the huge volume of data available. Here, the model cannot access the entire dataset simultaneously. Instead, data is accessed incrementally in chunks or batches, effectively simulating a streaming process where data points arrive over time.

The most studied case in literature is the classification problem. In this case, the data stream can be defined as an ordered and unbdoudend sequence of data points \mbox{$DS: d_1, d_2, ..., d_t, d_{t+1}, ...$}, where $d_t$ is the data point at time $t$, which is associated with a feature vector $X_t$ and a label $y_t$ (which is initially unknown). The goal is to predict the label $\hat{y}_t$ given the feature vector $X_t$, and, thus, learning the decision boundary defined by the probability distribution $P(y_t|X_t)$. The data stream results from a process called \textbf{concept}~\cite{cit:streaming_ts_read} which generates data points according to a probability distribution $P(X_t,y_t)$. Notably, this process is hidden and inaccessible.

The crucial problem that makes the traditional machine learning paradigm inapplicable is that $P(X_t,y_t)$ may change over time. As a result, the model learned using the previous concept becomes obsolete and requires updates or changes. In particular, the \textbf{concept drift} is the need for a change in the learned model due to a change in the concept~\cite{cit:cd_tsymbal2004,cit:cd_gama2019,cit:cd_delany2005,cit:cd_hoens2012}. Usually, these changes are unpredictable since the concept is hidden.

The literature distinguishes between virtual drifts and real drifts when considering the impact on the distribution $P(X_t, y_t)$. A real drift is defined as a change in the decision boundary $P(y_t \mid X_t)$~\cite{cit:cd_gama2014}. Conversely, the definition of a \textbf{virtual drift} has received many interpretations over the years. It was originally defined as a change in the input distribution $P(X_t)$ that necessitates a model update~\cite{cit:cd_tsymbal2004,cit:cd_delany2005,cit:cd_hoens2012,cit:cd_widmer1993}. Later works~\cite{cit:cd_gama2014,cit:cd_ditzler2015,cit:cd_khamassi2018,cit:cd_gama2019,cit:cd_wares2019} described it as a change in $P(y_t)$ or $P(X_t \mid y_t)$ that does not affect $P(y_t \mid X_t)$. However, a necessary definition must be explicitly made to capture whether the new concept directly undermines or merely extends the observed classification problem. Although introducing a new $P(X_t)$ or new classification labels may theoretically induce a change in $P(y_t \mid X_t)$, our distinction focuses on whether the classification rules for the input space observed so far remain unchanged. In this work, we refer to a virtual drift as a change in $P(X_t)$ that exposes the model to a new input space, potentially a previously unseen subspace or an extension of an existing one. The crucial aspect is that it preserves the decision boundary associated with all instances in the previously observed input space. This means that all past instances retain their original classification after the drift. Such a drift may even introduce new labels, as long as it does not alter the classification of earlier instances. Virtual drift requires the model to update its representation to generalize in new regions of the input space, without contradicting previously acquired knowledge. Conversely, we consider a \textbf{real drift} as a drift that changes the decision boundary associated with the input distribution observed in earlier concepts. In this case, the same instance may be classified differently before and after the drift. A real drift, therefore, modifies the relationship between features and target outputs, effectively changing the nature of the problem and invalidating past knowledge. In some cases, virtual and real drift may occur simultaneously, introducing both additional information in a new feature space and a transformation of the underlying problem associated with a feature subspace encountered in earlier concepts.

\begin{figure}[htbp]
    \centering
    \includegraphics[width=0.5\linewidth]{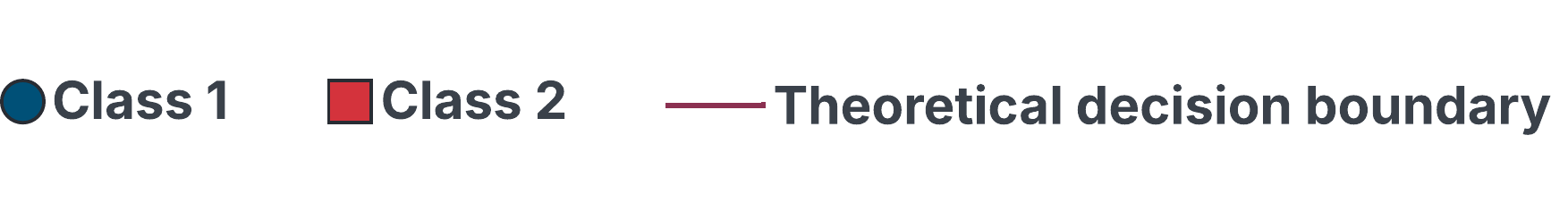}

    \begin{subfigure}[b]{0.4\textwidth}
        \centering
        \includegraphics[width=\textwidth]{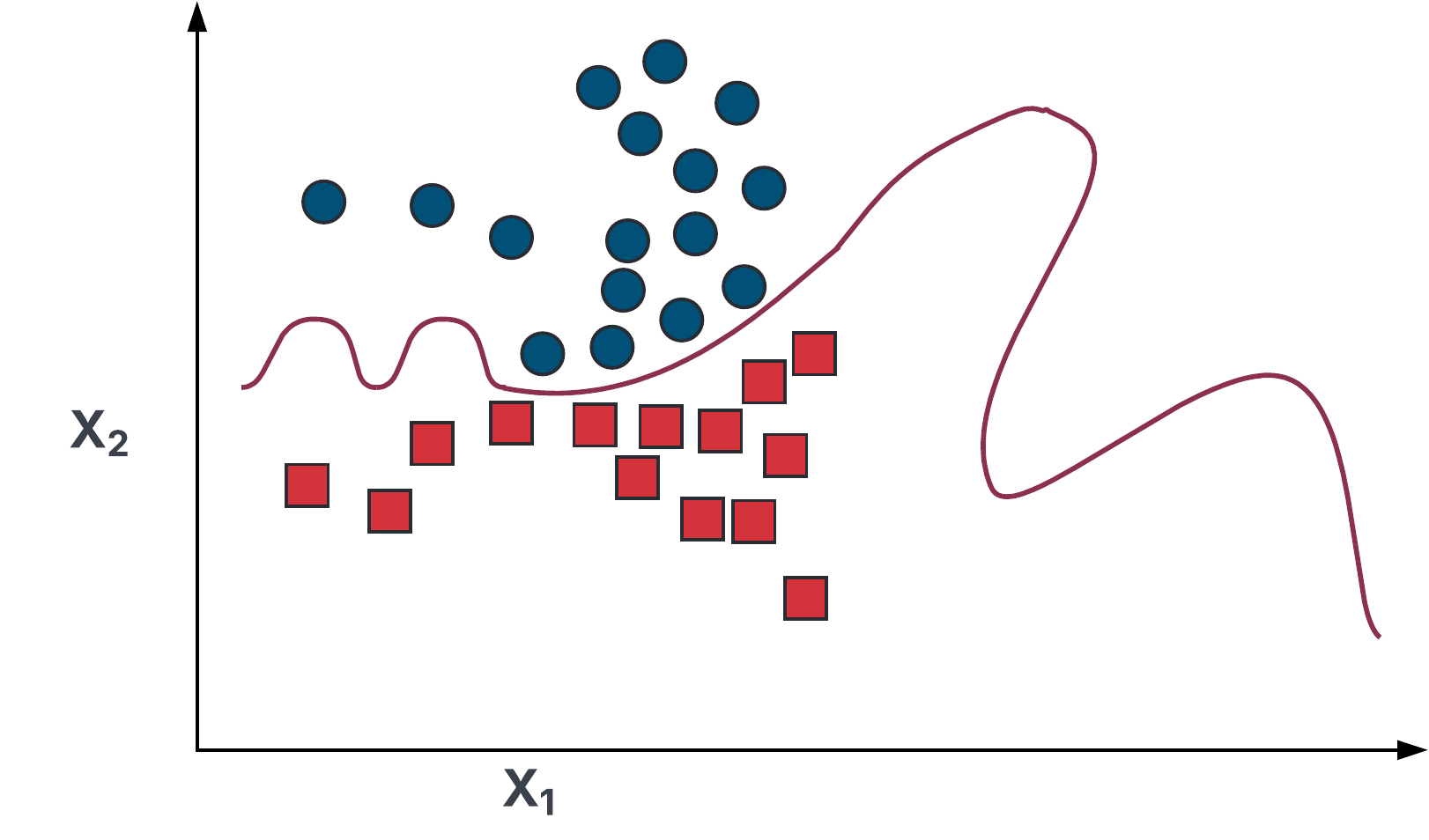}
        \caption{Before concept drift.}
        \label{fig:virtual_original}
    \end{subfigure}
    \hfill
    \begin{subfigure}[b]{0.4\textwidth}
        \centering
        \includegraphics[width=\textwidth]{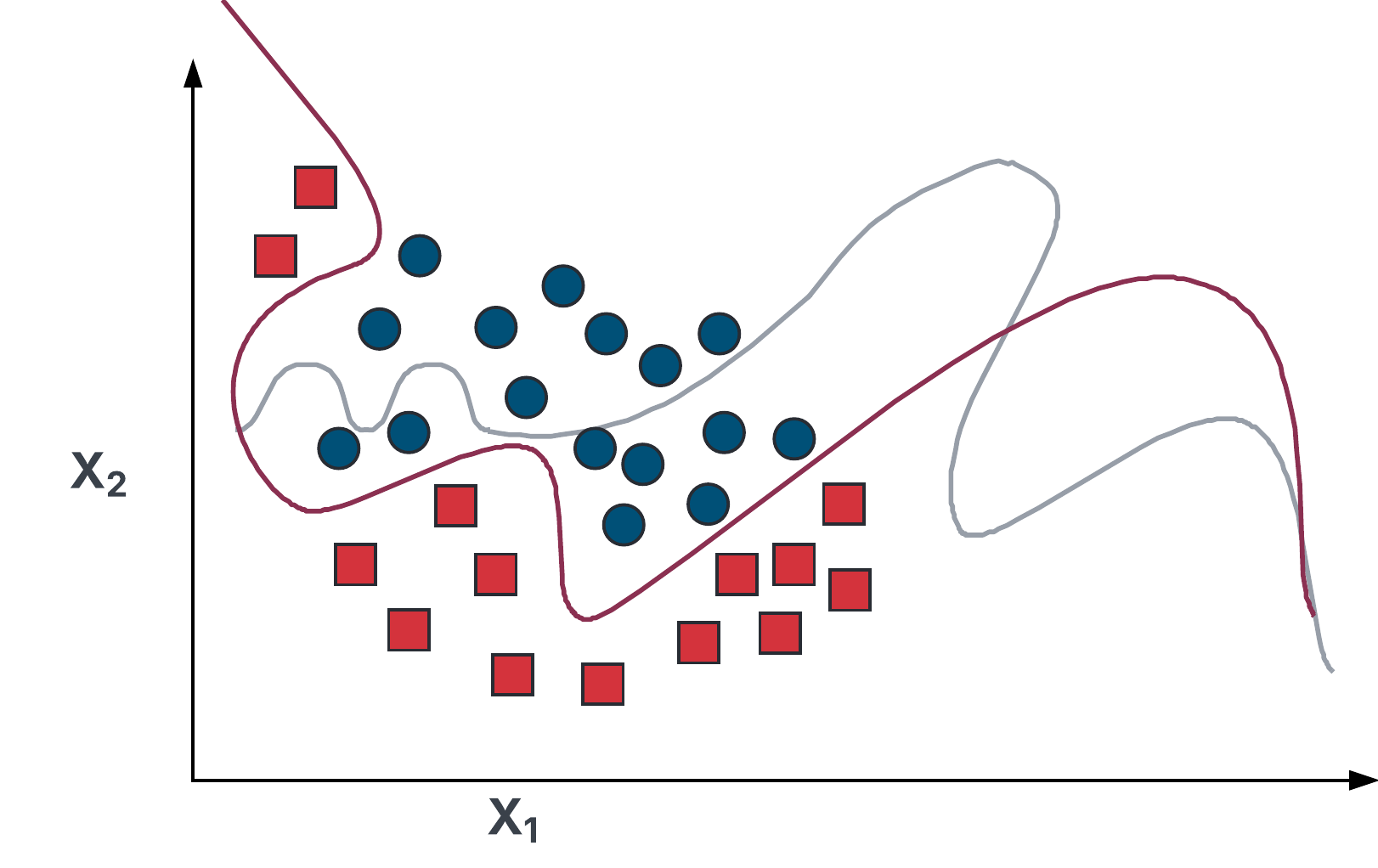}
        \caption{Real drift.}
        \label{fig:real_drift}
    \end{subfigure}

    \vspace{1em}

    \begin{subfigure}[b]{0.4\textwidth}
        \centering
        \includegraphics[width=\textwidth]{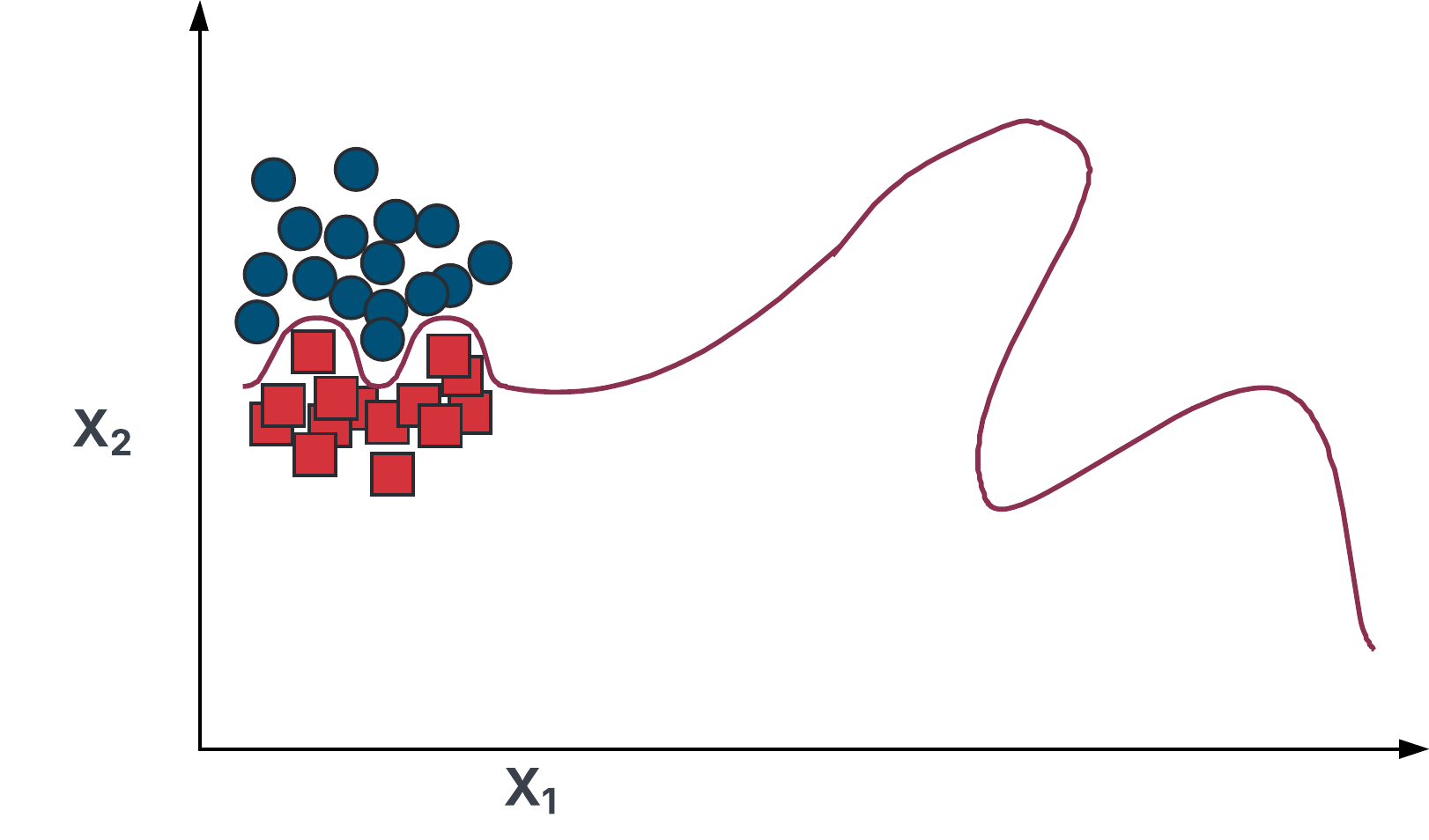}
        \caption{Zoom-in virtual drift.}
        \label{fig:virtual_zoom_in}
    \end{subfigure}
    \hfill
    \begin{subfigure}[b]{0.4\textwidth}
        \centering
        \includegraphics[width=\textwidth]{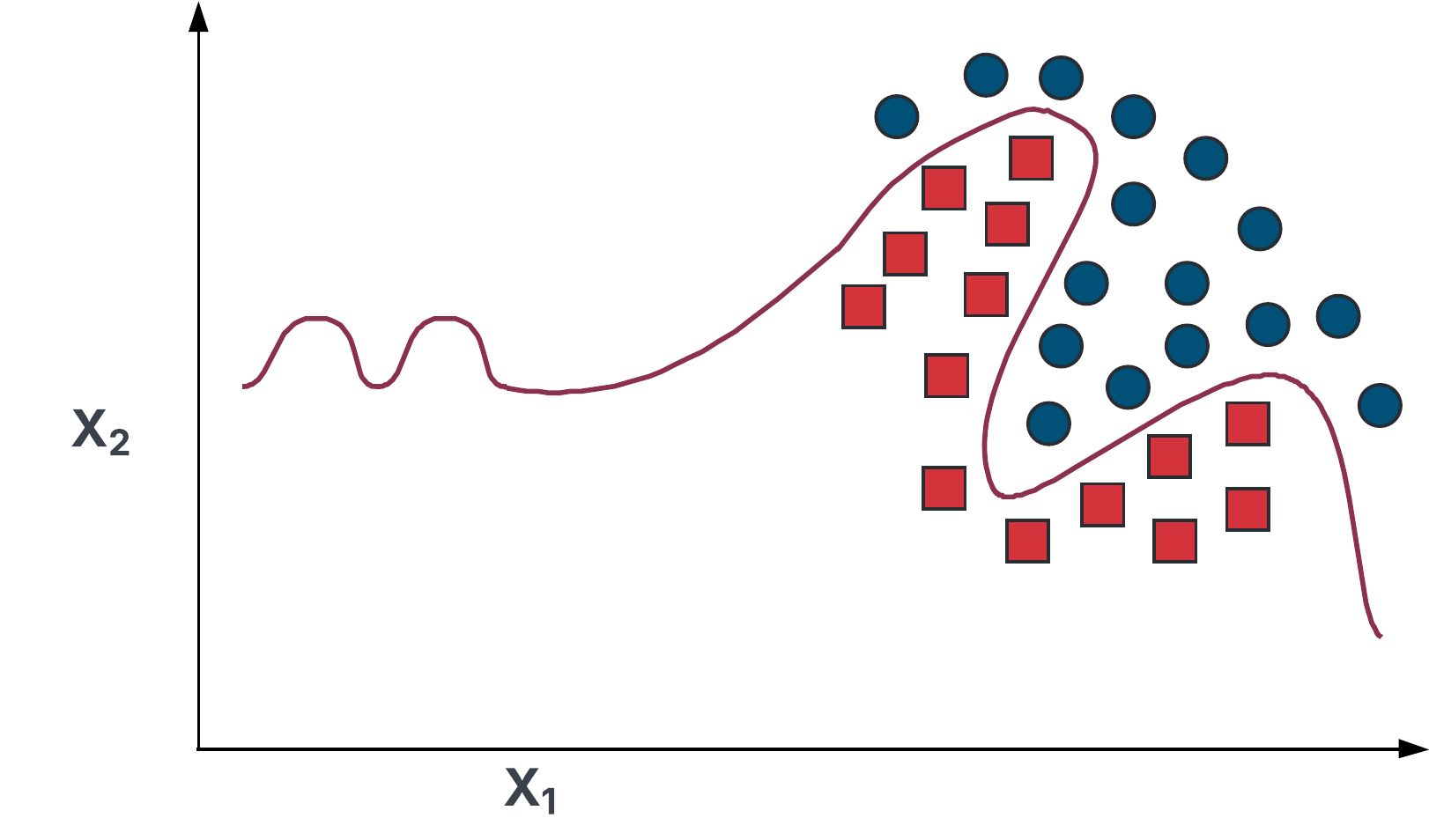}
        \caption{Pure expansionary virtual drift.}
        \label{fig:virtual_expansionary}
    \end{subfigure}

    \vspace{1em}

    \begin{subfigure}[b]{0.4\textwidth}
        \centering
        \includegraphics[width=\textwidth]{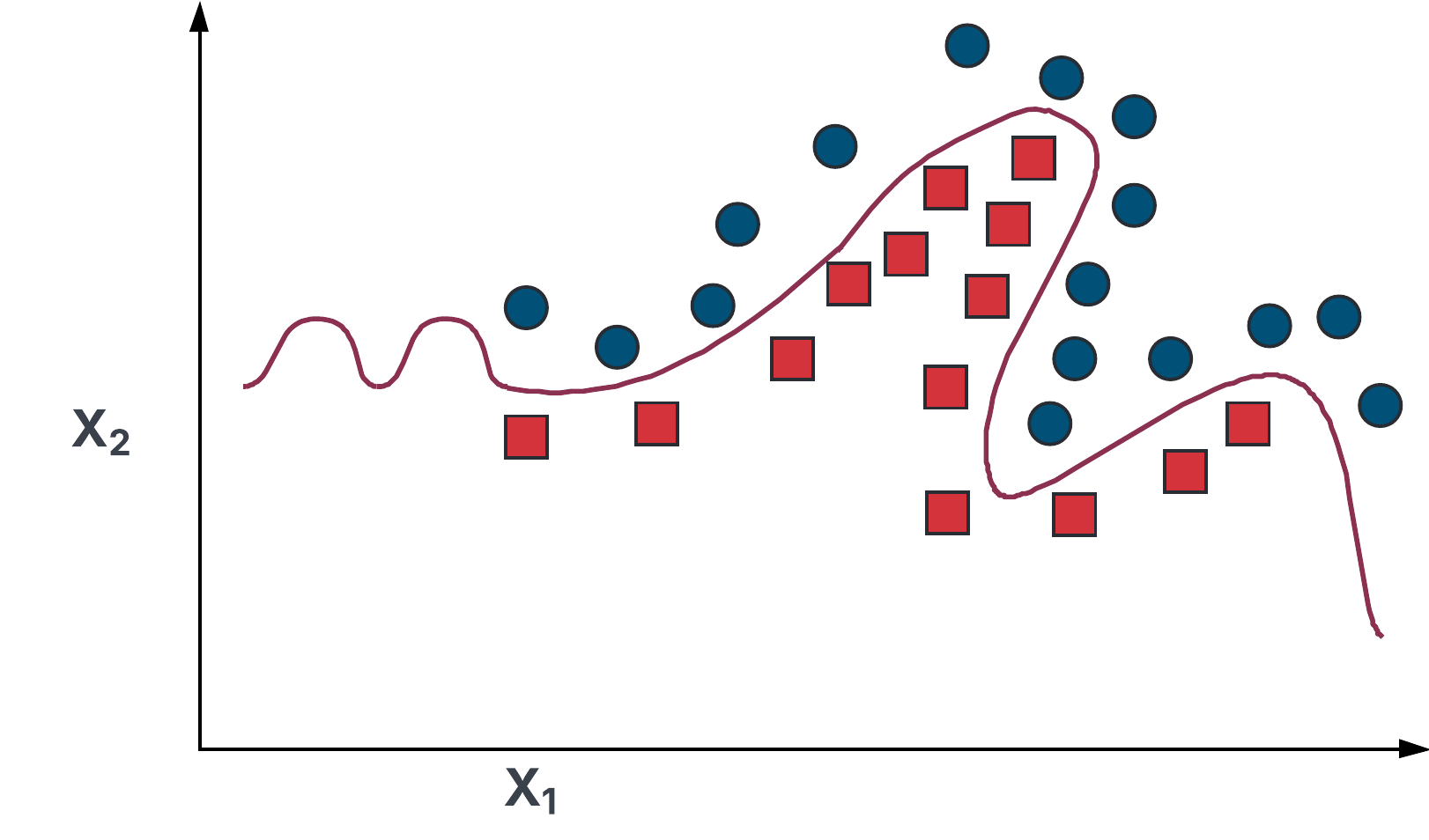}
        \caption{Expansionary virtual drift without real drift.}
        \label{fig:virtual_mixed_expansionary}
    \end{subfigure}
    \hfill
    \begin{subfigure}[b]{0.4\textwidth}
        \centering
        \includegraphics[width=\textwidth]{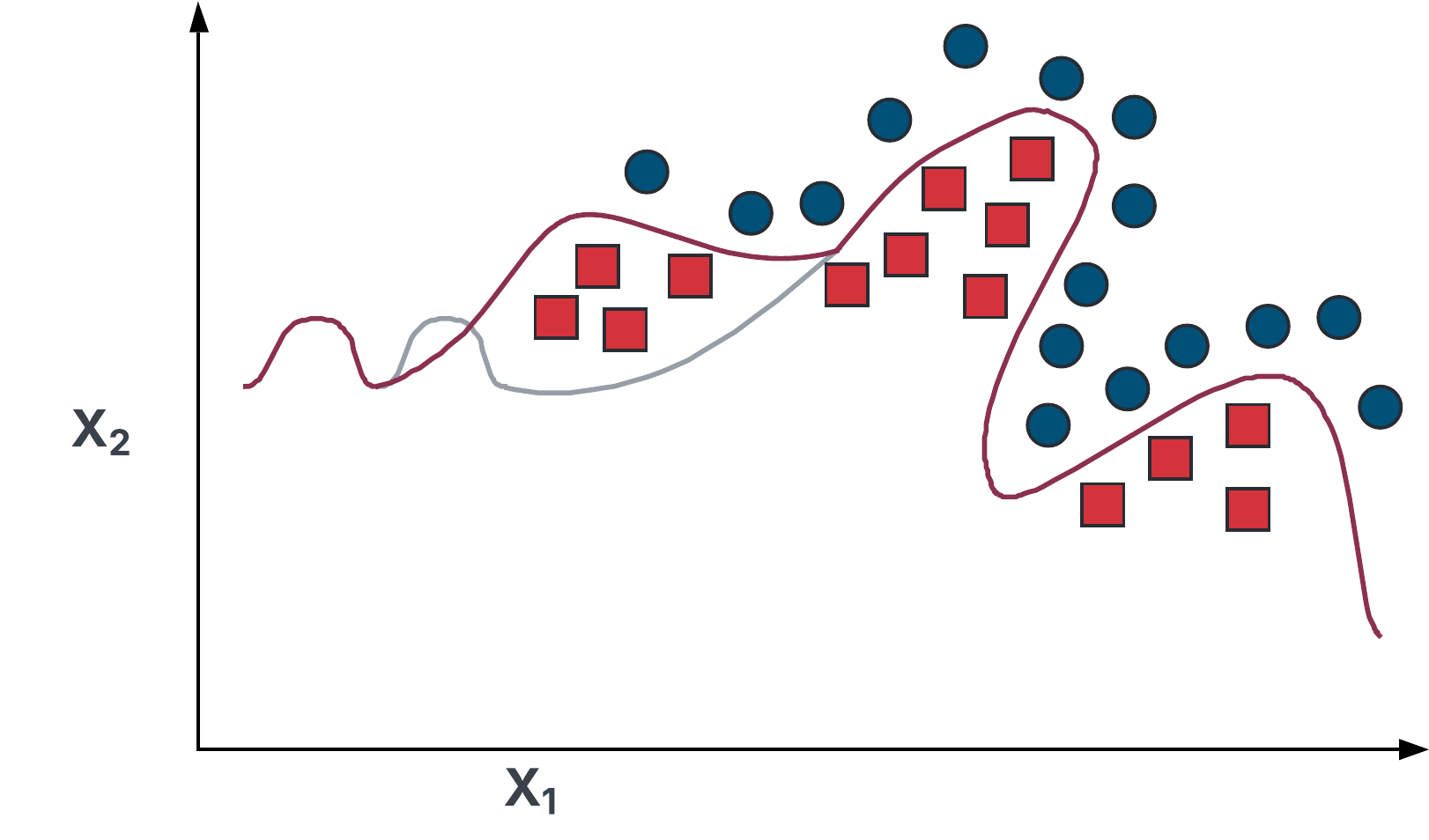}
        \caption{Expansionary virtual drift with real drift.}
        \label{fig:virtual_mixed_real}
    \end{subfigure}

    \caption{Illustrations of various concept drift types. The red line represents the theoretical decision boundary of the problem that the model must fit. \ref{fig:real_drift} depicts a real drift where the decision boundary shifts. \ref{fig:virtual_zoom_in} represents a \emph{zoom-in} virtual drift that concentrates on a subset of the initial feature space. \ref{fig:virtual_expansionary} shows a \emph{pure expansionary} virtual drift introducing a new, previously unseen feature subspace. \ref{fig:virtual_mixed_expansionary} displays a drift involving both familiar and new subspaces without altering the original decision boundary. Finally, \ref{fig:virtual_mixed_real} illustrates a combination of virtual and real drift, resulting in changes to the decision boundary within the previously known subspace.}

    \label{fig:drifts}
\end{figure}

Notably, virtual drifts are common, though not exclusive, in artificial streaming settings. While these settings do not involve data generated in real time, shifts in the input distribution $P(X_t)$ can still occur due to the order in which data is accessed or processed. For instance, large datasets read sequentially in batches may exhibit internal structure that leads to gradual distributional changes, even if the underlying task remains the same~\cite{cit:cd_widmer1993}.

Virtual drifts are often treated superficially in the existing literature. The nuances and implications of virtual drift have received relatively little attention. This work provides, for the first time, a more detailed analysis of virtual drift. We propose a categorization that distinguishes between different types of virtual drift, depending on how the input distribution changes and how those changes affect model performance. This finer-grained perspective allows for a better understanding of the learning dynamics involved and opens the door to a better understanding of SML and CL settings. \cref{fig:drifts} shows our categorization of concept drifts. For simplicity, we consider the case when the drift does not introduce new classes, but the problem is always defined on the same label set during the whole data stream. The red line represents the theoretical decision boundary of the problem, which the model is expected to learn. Note that this boundary is not known a priori. The model must infer it from the observed data.
  \cref{fig:virtual_original} represents the original distribution and \cref{fig:real_drift} depicts a real drift, which changes the boundary on the previously observed feature subspace. The remaining categories are described below.

A change in the input distribution $P(X_t)$ can, in some cases, reflect a shift toward a specific region of the feature space that was already present during the previous concept, though only sparsely represented. We refer to this scenario as \textbf{zoom-in virtual drift} (\cref{fig:virtual_zoom_in}). In such cases, the model trained on the earlier concept may underperform because it learned only a rough approximation of the decision boundary within that region, due to limited exposure to data points from it. After the drift, the input distribution becomes concentrated in this previously underrepresented subspace, revealing that the earlier model fails to accurately represent the true decision boundary there. As a result, the model must be updated, even if the actual underlying function has not changed.

Another notable case of virtual drift occurs when the input distribution $P(X_t)$ shifts to include a completely new region of the feature space that was not previously observed. We define this as \textbf{expansionary virtual drift}. In such scenarios, the existing model is unlikely to perform well, as it has no prior experience with data from this newly introduced subspace. Expansionary virtual drift can be \emph{pure} (\cref{fig:virtual_expansionary}), where the new concept generates data solely in the novel subspace. Alternatively, the new concept may include data from both the previously seen and newly introduced subspaces. In this mixed case, the new concept may also alter the decision boundary within a portion of the original feature space, resulting in the coexistence of both virtual and real drift (\cref{fig:virtual_mixed_real}).

Along with the type of change, \textbf{severity} is an important marker for quantifying how the concept has changed with respect to the previous one~\cite{cit:concept_drift_descriptors,cit:cd_gama2019,cit:cd_webb2016}. Formally, severity measures the discrepancy between the joint distributions $P(X_t,y_t)$ of features and labels before and after the drift.  Associated with the severity, the \textbf{influence zone} of the drift indicates whether the change affects a limited region or the entire feature space~\cite{cit:concept_drift_descriptors,cit:cd_gama2019}. In the case of real drifts, analyzing the influence zone can help understand what has changed and must be relearned. A low-severity drift, or \emph{mild} drift, will probably be associated with a limited influence zone, thus requiring updating the learned decision boundary with respect to a narrow region of the input space. Conversely, \emph{severe} real drifts may change the decision boundary associated with a significant region of the input space, thus demanding substantial relearning and forgetting of past knowledge. 

Additionally, concept drifts vary in their \textbf{speed of occurrence}~\cite{cit:cd_gama2014,cit:cd_gama2019}. \emph{Abrupt drifts} happen suddenly, causing immediate shifts in the data distribution. \emph{Gradual drifts} involve a slow transition with overlap between old and new concepts, whereas \emph{incremental drifts} proceed through intermediate stages of gradual change. Finally, \emph{recurring drifts} occur when previously seen concepts reappear, requiring the model to efficiently recall past knowledge.

\section{Streaming Machine Learning}\label{sec:sml}
SML~\cite{book_bifet} is usually concerned with natural streams, often involving real-time analytics where data is continuously collected over time. In this situation, the emphasis is on continuously training a model and quickly adapting to changes as new data points arrive. These scenarios often require several strict constraints. Each data point is processed only once, as only a single pass over the stream is permitted. To maintain responsiveness, the computational time required to process each instance must be minimal. Similarly, memory usage is constrained and should grow at most sublinearly with respect to the total length of the data stream. Moreover, the model is expected to be available for prediction at any point during the stream. 

The primary objective of an SML system is to promptly detect and adapt to concept drifts (both virtual and real). Such systems focus exclusively on the current data distribution and the active concept, aiming to provide accurate predictions under present conditions. The main challenge lies in minimizing the adaptation time following a drift, as the model's outputs during this period are typically unreliable. Consequently, knowledge from past concepts is often discarded once a drift occurs and is no longer used to support future predictions.

\paragraph{SML Setting and Evaluation} A data stream is defined by SML literature as an ordered and unbounded sequence of data points $\mathcal{S} = d_1, d_2, ..., d_t, d_{t+1}, ...$, where $t$ is the timestamp. As already mentioned in Section~\ref{sec:non_stationary_ds}, in the classification problem, each data point $d_t$ is represented by a feature vector $X_t$ and a label $y_t$. SML assumes that, at timestamp $t$, the model receives the feature vector $X_t$ and must predict the label $\hat{y}_t$. Before receiving the new feature vector $X_{t+1}$, the real label $y_t$ is given, and the model can update its performance score using the pair $(X_t,y_t)$ and learn from it. This paradigm is referred to as \textbf{prequential evaluation} or test-then-train evaluation~\cite{cit:prequential,cit:prequential2013}.

\paragraph{Drift Detectors} The design of drift detectors has attracted substantial interest within the SML community. Gama et al.~\cite{cit:cd_gama2019} propose a taxonomy of the main approaches, broadly distinguishing three families. \emph{Error rate-based methods}, such as ADWIN~\cite{cit:adwin}, the Early Drift Detection Method~\cite{cit:eddm}, EWMA~\cite{cit:ewma}, and EWMA for Concept Drift Detection~\cite{cit:ecdd}, track the online classification errors of base learners to identify noteworthy deviations. These techniques are particularly suited to real and contradictory drifts, where changes in classification rules lead to performance degradation. \emph{Data distribution-based} approaches assess discrepancies between historical and incoming data, making them appropriate for detecting input or virtual drift. Notable examples include Concept Drift via Competence Models (CM)~\cite{cit:cm}, as well as techniques relying on information-theoretic measures~\cite{cit:det_data1} and statistical bounds on data streams~\cite{cit:det_data2}. \emph{Multiple hypothesis testing} strategies employ sets of statistical tests to capture drifts, as exemplified by Just-in-Time (JIT) adaptive classifiers~\cite{cit:det_hyp1}, hierarchical three-layer schemes~\cite{cit:det_hyp2}, and lightweight detection ensembles~\cite{cit:det_hyp3}. Selecting an appropriate drift detector is therefore a key challenge in the streaming context.


\paragraph{SML Algorithms} SML classification algorithms can generally be categorized into five main methodological families, each adopting distinct mechanisms to handle the constraints of data streams~\cite{cit:sml_strategies}. They usually start from offline machine learning approaches and produce online and streaming versions able to learn continuously from an unbounded data stream and adapt to changes.

Frequency-based approaches estimate posterior class probabilities by incrementally collecting statistics on the frequency of features. These methods assume conditional independence among attributes given the class label and typically employ Bayes' theorem to perform predictions. A canonical example is the Naïve Bayes (NB) classifier~\cite{cit:nb}, including several revised versions designed to improve its performance in streaming contexts, like Sketch-Based NB~\cite{cit:nb_streaming}, Ageing-Based Multinomial NB~\cite{cit:mnb_streaming}, and Incremental Weighted NB~\cite{cit:incr_weighted_nb}.

Neighborhood-based methods, such as variants of the k-Nearest Neighbors (KNN) algorithm, base their predictions on proximity in the feature space. To make this computationally feasible in a streaming setting, they maintain a sliding window of recent samples rather than storing the entire history. Notable examples are KNN with Self Adjusting Memory~\cite{cit:knn_streaming} and Efficient KNN Graph Construction~\cite{cit:efficient_knn}.

Tree-based strategies adapt classic decision tree algorithms for incremental learning. These methods typically employ the Hoeffding bound~\cite{cit:hoeffding_bound} to select split attributes with high statistical confidence from streaming data. Early implementations~\cite{cit:vfdt} do not handle concept drifts effectively, whereas later extensions such as the Hoeffding Adaptive Tree (HAT)~\cite{cit:hat} incorporate change detection mechanisms like ADWIN at each node to dynamically adjust the tree structure.

Ensemble-based classification represents another prominent category. These models combine the predictions of multiple base learners, often trained on resampled subsets of the data~\cite{cit:online_bagging,cit:leverage_bagging,cit:streaming_random_patches}. One of the most well-known solutions is the streaming version of Random Forest called Adaptive Random Forests (ARF)~\cite{cit:arf}. By introducing diversity among learners, ensemble techniques aim to improve generalization performance~\cite{cit:ensemble}.

Neural network-based approaches have started to receive increasing attention due to their capacity for high-level abstraction. However, their application in streaming scenarios is still poorly investigated since they present challenges related to real-time learning, memory efficiency, and parameter tuning. Notable examples are Continuously Adaptive Neural Networks for Data Streams~\cite{cit:streaming_hyper_parameters_tuning_gunasekara}, Autonoumous Deep Learning~\cite{cit:autonomous_deep_learning}, Incremental LSTM~\cite{cit:ilstm}, Continuous LSTM~\cite{cit:cpnn,cit:tenet}. 

\paragraph{SML Libraries} The landscape of SML libraries has evolved from traditionally Java-centered ecosystems toward increasingly Python-oriented solutions, reflecting the changing needs of both researchers and practitioners. \emph{MOA (Massive Online Analysis)}\footnote{\url{https://moa.cms.waikato.ac.nz/}}~\cite{book_bifet} remains the reference library for SML in Java. Building on the WEKA project, MOA provides a mature and comprehensive framework with a wide range of algorithms for online classification, clustering, and outlier detection, as well as standardized evaluation protocols that ensure experimental rigor and reproducibility.

In the Python ecosystem, River\footnote{\url{https://riverml.xyz/dev/}}~\cite{cit:river} initially emerged as the main reference library for incremental learning on data streams, following the merger of Creme and scikit-multiflow. River adopts a native streaming paradigm in which data are processed one instance at a time and emphasizes flexibility through lightweight, dictionary-based data structures that support evolving feature spaces.

CapyMOA\footnote{\url{https://capymoa.org/}}~\cite{cit:capymoa}
bridges these two ecosystems by exposing MOA's robust Java-based algorithms through a Python interface. By directly leveraging MOA's highly optimized implementations, CapyMOA achieves substantially higher efficiency than fully Python-native libraries such as River, particularly in large-scale or high-throughput scenarios. Combined with seamless integration into the Python machine learning ecosystem (e.g., PyTorch and scikit-learn) and support for advanced settings such as complex concept drift simulation, semi-supervised learning, and online continual learning, CapyMOA is emerging as a strong candidate reference library for SML in Python.

\section{Continual Learning}\label{sec:cl}
Similar to SML, CL also studies data streams subject to concept drift. Unlike SML, CL focuses more on knowledge retention rather than quick adaptation. Historically, this is due to the first empirical observation of the \emph{catastrophic forgetting} phenomenon in artificial neural networks \cite{french1991, robins1995}. Forgetting is exacerbated on data streams with sudden, virtual drifts, where new concepts replace previous ones \cite{masana2023}. Given the large degree of plasticity in artificial neural networks trained with backpropagation and gradient descent, whenever a concept is not observed for a sustained period, it is quickly forgotten to learn new concepts. \\
Since forgetting is more crucial in the presence of virtual drifts, CL originally started by focusing on data streams where concepts never overlap, and come one after the other \cite{rebuffi2017}. This aligns well with the artificial stream perspective, where a huge amount of data is scanned sequentially and, when observing new distributions (due to the rise of virtual drifts), one aims to learn it without forgetting what has been learn in the past. This allowed researchers to design ad-hoc strategies to mitigate forgetting \cite{masana2023, verwimp2024}. \\
The \emph{class-incremental} scenario is one of the first and most studied CL scenarios \cite{rebuffi2017}. The learning model, usually an artificial neural network, is required to i) solve a classification task where ii) classes are presented sequentially one after the other, and iii) once a class disappears after a drift, it never occurs again (no recurring drifts). Class-incremental scenarios essentially partition a given data stream into subsets with non-overlapping classes.\\
Class-incremental scenarios can be easily derived from any classification dataset. For example, one can take the MNIST dataset composed of images of digits, and build a class-incremental scenario by considering only pairs of digits at once. For instance, the model is first trained on all 0 and 1 digits (the first ``task''), then on all 2 and 3 digits, and so on up until all 8 and 9 digits. At any point during training, the network should be able to correctly classify all digits seen up to that point. However, due to forgetting, a simple fine-tuning of the model leads to almost zero accuracy on previous digits, as the model only predicts the most recent digit pair.\\
Sometimes, a class-incremental scenario also provides task labels during inference, to help the model understand which task the data comes from. This scenario allows building modular architectures with task-specific components. It is therefore called task-incremental learning \cite{vandeven2018}.\\
If class-incremental is the most challenging scenario for forgetting, it is by no means the only CL scenario. One can easily introduce repetitions of previous concepts (i.e., classes) in the data stream. Alternatively, one can only introduce new examples of existing classes. This scenario is usually called \emph{domain-incremental} \cite{vandeven2018}. Continuing with our example based on MNIST, a domain-incremental scenario might require the network to classify digits as either even or odd. If the data stream presents examples of two classes at a time, the model has to continuously adjust its idea of what an even or odd digit is. This scenario is usually associated with a lower amount of forgetting, as it does not introduce completely new concepts. \\
Notably, all the aforementioned scenarios present virtual drift, since each new experience introduces a new input distribution that was not observed during the past. Particularly, the situation presented in Figures~\ref{fig:virtual_zoom_in} and \ref{fig:virtual_expansionary} explicitly maps the domain-incremental setting. The class-incremental setting, in addition to introducing a new input distribution, also complicates the decision boundary by adding new classes in the new input distributions.  CL rarely studies streams with real drifts. Such streams would require not only to forget outdated information, but also to \emph{selectively} forget knowledge, a challenging endeavour with deep neural networks, where representations are usually entangled. \\

Given the type of scenarios and data streams faced by a CL model, researchers derived different ideas on how to combat forgetting. While a comprehensive overview of CL approaches is beyond the scope of our work, we summarize here some of the most important classes of approaches. This will also be useful when introducing a motivating example for Streaming Continual Learning, later in Section \ref{sec:example}.\\
As we have seen, a class-incremental scenario presents new concepts sequentially without repetition of previous ones. This breaks the \emph{i.i.d.} assumption, which many machine learning approaches rely on. To better approximate an i.i.d. distribution, \emph{replay strategies} \cite{hayes2021} keep a storage of previously seen examples (e.g., K examples per class). At each training iteration, the model receives the current examples from the stream and a sample from the replay memory. Intuitively, replay mitigates forgetting as it restores knowledge from previous concepts. However, CL requirements usually do not allow for storing the entire data stream in memory, thus effectively limiting the replay memory size. Recent works explore settings with unlimited memory but limited computational budget, as they model some real-world cases where storage cost is often negligible.\\
Replay is by far the most popular CL approach for class-incremental scenarios. Regularization approaches \cite{parisi2019} are also among the most studied ones. The idea is to penalize changes in the current model that would drive it away from the optimum for previous concepts. This high-level idea is instantiated differently depending on the context: i) each weight can bring an importance value related to previous tasks and receive an update inversely proportional to its importance; ii) the current model might be encouraged to approximate its predictions on previous tasks through distillation, to remain functionally similar to the previously reached optimum; iii) the update direction might be constrained to remain (exactly or approximately) orthogonal to previous update directions to mitigate interference. Other classes of CL approaches, called \emph{architectural strategies}, dynamically expand the model upon receiving new data, usually leading to multi-module models with task-specific components \cite{rusu2016, sokar2021}. Naturally, different CL approaches can be combined to take the best of several possible options (hybrid approaches). \\

\paragraph{Evaluation in CL} Unlike SML, prequential evaluation is rarely used in CL. The CL framework includes a stream $\mathcal{S} = (s_1, s_2, \ldots)$ of a potentially infinite amount of experiences (or tasks, or concepts). Each $s_i$ provides a set of examples, and drifts usually occur between any experiences. The model is not subject to strong constraints on the allotted training time on each experience, and it can therefore be trained for multiple epochs until convergence. At any point in time, the model is evaluated on a set of unseen examples (the test set), which are held out from the data in each experience. When training on $s_i$, the evaluation on the test set of $s_i$ measures the generalization performance of the model. Differently, when training on $s_i$, evaluating on the test set of $s_j, j<i$ measures the ability of the model to retain previous knowledge. That is, the forgetting. More formally, let us consider the supervised classification task (a similar reasoning applies to other tasks, like regression). The evaluation metric is the prediction accuracy (percentage of correctly classified examples). The metric ACC \cite{lopez-paz2017} is computed as $\frac{\sum_i a_{i, N}}{N}$, where $N$ is the total number of experiences encountered by the CL agent and $a_{i,j}$ is the average accuracy on the test set of $s_i$ after training on $s_j$. ACC measures the overall predictive performance of a CL agent on a set of $N$ experiences. \\
The Backward Transfer (BWT) \cite{lopez-paz2017} measures the knowledge retention. It is therefore equal to negative forgetting ($BWT = -F$ with $F$ forgetting). For a given experience $s_i$, its BWT is computed as $a_{i, j} - a_{i, i}, j>i$. The first term measures the performance on experience $i$ after training on the (current) task $j$, while the second term measures the original performance on experience $i$ right after training on it (it is the reference value). A positive value indicates that learning new experiences benefits the performance on $s_i$. On the contrary, a negative value denotes forgetting. When averaged over a set of tasks, the forgetting metric is called Average Forgetting. \\
This short overview of evaluation in CL highlights how the model always has access to a set of held-out examples from each experience, even though they are only used for evaluation and never for training.

\paragraph{Online continual learning (OCL)} As we have seen, CL does not impose strict constraints on the allotted training time or on the number of examples the model accesses at a time. We call the setup presented so far ``batch CL'', to highlight that each experience carries a large amount of examples, and the model can be trained over multiple epochs until convergence. In contrast, OCL \cite{soutif-cormerais2023} focuses on a scenario much more similar to SML. Data arrives one ``minibatch'' at a time. In practice, this means that each experience carries only a few examples. The model can be trained for a few iterations on the same minibatch. However, this setup is much more restrictive than a multi-epoch training on a large batch of data. OCL preserves the evaluation setup where the model can be queried at any time on data belonging to previous experiences. One popular evaluation metric in OCL requires computing the predictive accuracy at any point, even after each iteration. The average of these measurements over the entire stream is called the Average Anytime Accuracy \cite{caccia2022}.\\ 
The CL scenarios presented before still apply. For example, the class-incremental scenario can be easily repurposed to OCL by splitting each experience into smaller experiences with only a few examples. As a result, OCL streams are composed of more experiences than CL ones.\\
Even though some of the challenges of SML and CL meet in the OCL scenario, the current literature is still dominated by the CL view: learning models are usually deep artificial neural networks, mitigating forgetting remains the main focus, and virtual drift is the most common type of drift.\\
However, OCL highlights the need for fast adaptation to new knowledge, as new data points arrive quickly, and training cannot be performed over multiple epochs. In addition to these strict time constraints, OCL also obeys the memory constraints of CL: no storage of the entire history of the data stream is allowed. Replay is still used in OCL, by leveraging a memory buffer of \emph{fixed-length}. All these characteristics will be shared by the SCL framework.

\paragraph{Continual Reinforcement Learning}\label{sec:rl}
\begin{figure}[ht]
\centering
\includegraphics[width=0.5\textwidth]{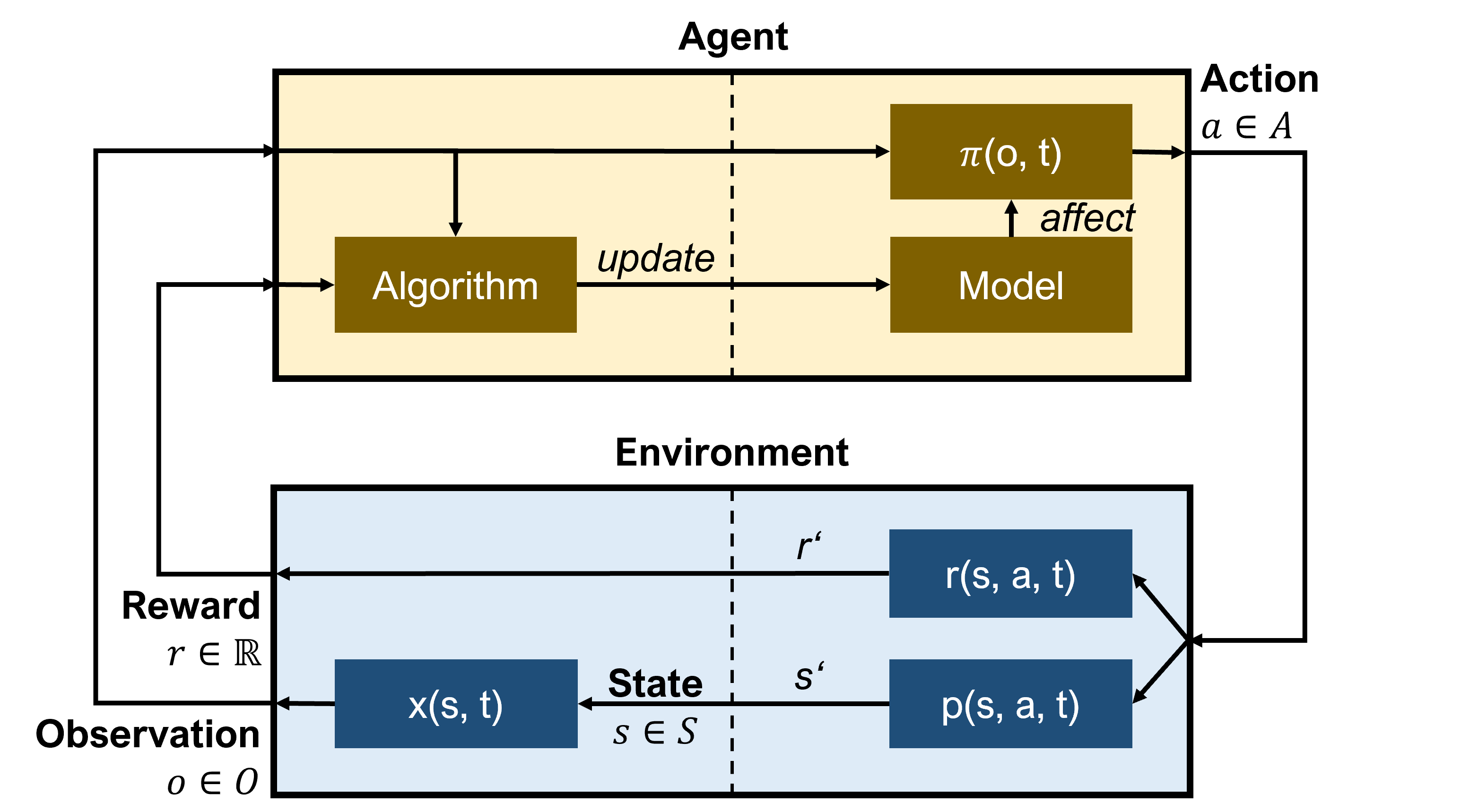}
\caption{\label{fig:rl} An RL control loop: as a reaction to actions $a(t)$ taken by the agent, it receives observations $o(t)$ and rewards $r(t)$. }
\end{figure}

In Reinforcement Learning (RL), an agent implements a continuous, reciprocal relationship with the environment it is situated in: the agent continuously takes actions, which impact the environment and thus the observations and rewards obtained from it, see \cref{fig:rl}.
This involves learning from non-stationary data distributions even when environments themselves can be considered stationary. This is mainly because the assessment of different actions will evolve as learning progresses, leading to different actions being taken in similar situations, and thus to different observations.
As such, this would mainly represent virtual drift (observations change, but not the optimality of actions in the same circumstances). 

\begin{figure}
\centering
    \includegraphics[width=0.7\textwidth]{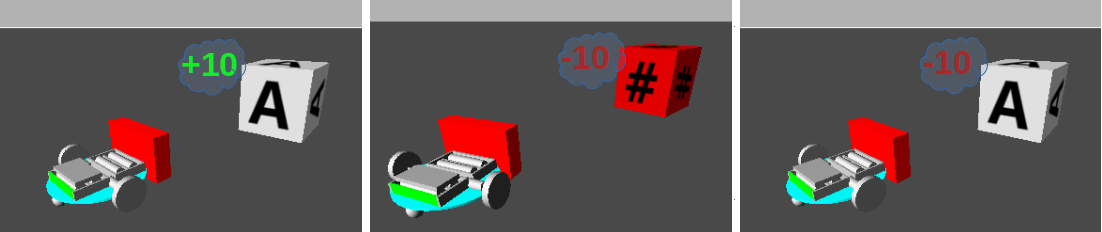}
    \caption{\label{fig:crl} Virtual and real drift in Continual Reinforcement Learning (CRL). In this example scenario, a robot is rewarded or punished when touching certain objects, depending on form, color, and displayed symbol. Left: an object that should be touched, giving +10 reward. Middle: virtual drift, introducing a different object that should not be touched (or else: +10 reward). Right: real drift, changing the reward previously obtained for touching the object to punishment.
    }
\end{figure}

However, in Continual Reinforcement Learning (CRL), we consider environments themselves to be non-stationary. This implies that, for the same observation, different actions may be required for maximum reward at different points in time, which corresponds to a real shift, or even a combination of both (\cref{fig:crl}).

\section{Streaming Continual Learning}\label{sec:scl}
\begin{figure}[t]
    \centering
    \includegraphics[width=0.8\textwidth]{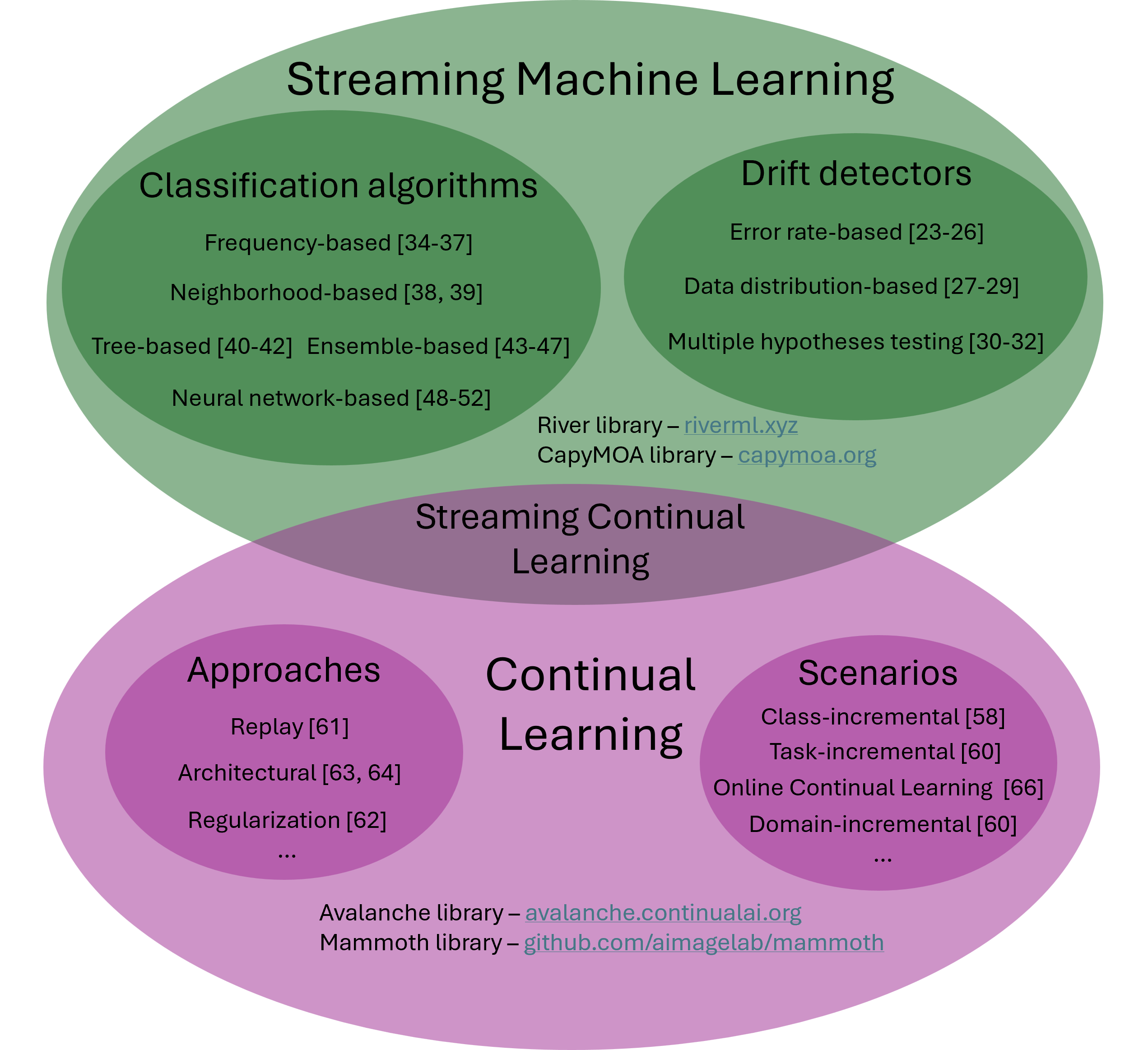}
    \caption{Summary of SML and CL, with some examples of popular state-of-the-art approaches and popular libraries. We refer to existing reviews for an in-depth treatment of either field.}
    \label{fig:summary}
\end{figure}
SCL emerges from the need to unify the complementary goals of SML and CL (summarized in Figure \ref{fig:summary}) when operating on evolving data streams, which may experience both real and virtual drifts.
SML focuses on the current concept and aims to perform well on the currently observed feature subspace. SML does not address the issue of forgetting and makes no effort to assess whether knowledge from previously observed subspaces remains valid. Its main objective is to quickly adapt to new concepts, even at the cost of losing prior information. Since both real and virtual drifts require model updates, SML literature has often claimed that distinguishing between them is of limited practical relevance~\cite{cit:cd_tsymbal2004,cit:cd_delany2005,cit:cd_zliobaite2010,cit:cd_hoens2012,cit:cd_wares2019}.\\
Conversely, CL typically assumes that concept drifts are only virtual (often expansionary), meaning that previously acquired knowledge remains valid. The goal in this case is to retain previously learned knowledge while incorporating new information.\\

This distinction is also reflected in the evaluation protocols used in each field. SML usually adopts prequential evaluation (Section~\ref{sec:sml}), assessing performance only on the current concept. This fits natural streaming scenarios, where data are generated in real time, and the focus is on immediate adaptation. In contrast, CL evaluates performance across all previously encountered concepts (Section~\ref{sec:cl}), which is more suited to artificial streaming scenarios, where the full problem cannot be observed at once. Instead, it is split into subproblems (CL experiences), and the model is expected to remember past knowledge while learning new tasks.

Blending these two views into a unified framework is at the core of SCL. From this perspective, the model should maintain a general understanding of the entire stream while actively relying only on knowledge that remains valid for the current concept, or on reinterpreting prior knowledge in a way that supports current learning. Any knowledge that is not immediately useful should be preserved and made accessible for potential future reuse.

Crucially, as discussed in~\cite{cit:scl}, the environment itself dictates which knowledge should be retained or discarded. New data may introduce patterns that either complement (virtual drifts) or replace existing ones (real drifts). In some situations, prior knowledge becomes irrelevant and should be suppressed to prevent interference. In others, it may still be valuable, either because it partially overlaps with new concepts or because it is likely to recur. Moreover, real drifts may not invalidate prior knowledge entirely: mild drifts or those with limited influence zones might still leave large portions of the input space unaffected.

Furthermore, in many real contexts, forgetting should not be seen solely as a drawback but as a necessary strategy for managing limited resources. The model should retain general, reusable knowledge while dynamically adapting to changes in the environment. This implies a more selective and context-aware form of forgetting, where information is not erased indiscriminately but evaluated based on its relevance to the ongoing data stream.

\paragraph{SCL and Online CL} As we mentioned in Section \ref{sec:cl}, research in OCL already took an important step in the SCL direction. Firstly, OCL considers data streams where only a few examples are available at a time. Secondly, OCL incorporates anytime evaluation at a high frequency (up to every training iteration). Thirdly, OCL shares the tight computational constraints of SCL. However, the current direction of OCL research is still heavily oriented towards the mitigation of catastrophic forgetting, where neural networks and class-incremental learning scenarios are prominent. SCL aims at tackling more general scenarios, with different types of drifts possibly appearing in the data stream (e.g., real drifts). SCL offers the opportunity to build a shared vocabulary, connecting (O)CL and SML researchers, expanding the scope of their application domains, and leveraging existing works to design hybrid approaches.

\paragraph{SCL approaches} SCL paradigms should exhibit five essential capabilities~\cite{cit:scl}: (1) rapid adaptation to both real and virtual drifts, (2) autonomous detection of concept drifts, (3) ability to learn effectively from single (or a few) data points, (4) capacity to form hierarchical, structured representations, and (5) selective retention of relevant knowledge to mitigate forgetting. Notably, the ability to build hierarchical representations contributes directly to better knowledge selection, as it enables the model to structure the knowledge and identify which components could be reused or adapted.

In this context, SML methodologies could play an important role in monitoring and detecting changes and quickly adapting to them, while learning continuously from single data points. On the other side, CL and OCL are explicitly meant to structure and organize latent representations using deep learning models and apply strategies to mitigate forgetting.

Notably, Replay and regularization CL strategies may show limitations in streaming settings characterized by real concept drifts. These approaches typically attempt to retain and jointly optimize for all previously seen concepts. However, when past and current concepts conflict (as is common with real drifts), this simultaneous optimization can introduce incompatible learning signals, ultimately preventing convergence.

In contrast, \textbf{architectural strategies} (another popular and varied set of CL approaches) offer a more modular and scalable alternative. By structurally decoupling the parameters associated with different concepts, these methods can mitigate interference and allow for selective reuse of relevant components. Such architectures enable the system to isolate, preserve, and repurpose knowledge through mechanisms like dynamic module expansion or masking mechanisms. This design allows the learner to focus adaptively on the current concept while retaining access to past knowledge, aligning closely with the core principles of SCL, where adaptability and scalability are crucial.

Regarding the evaluation procedure, one may apply both prequential evaluation to measure how the model is performing on the current concept and classical CL evaluation to estimate how the model retains the past concepts. Tracking how the model forgets over time provides valuable insight into the effects of newly acquired knowledge. Even when a previously learned concept loses its relevance, analyzing shifts in its accuracy can offer important information. Ideally, the model should preserve high-level knowledge that is common across concepts, as this can aid in solving current tasks~\cite{cit:scl}.

Following this direction, we presented a proposal for an architecture that can embody SCL solutions: \textbf{fast and slow learners}~\cite{cit:scl} and takes inspiration from the Complementary Learning Systems (CLS) theory~\cite{cit:cls1,cit:cls2}. The CLS theory distinguishes between a fast-learning system for rapid adaptation and a slow-learning system for stable, long-term consolidation. This duality naturally maps onto the SCL paradigm, where a fast learner (SML) quickly adapts to incoming data, while a slow learner (CL) retains generalizable knowledge and mitigates forgetting. The interaction between the two is bidirectional: the slow learner benefits from relevance signals and recurrence detected by the fast learner, while the fast learner can exploit structured representations built over time by the slow learner. This coordination enables SCL to adapt rapidly to change while preserving useful knowledge across the stream.

\paragraph{Datasets for SCL} As demonstrated by the quick rise of class-incremental scenarios in CL, leveraging existing datasets and benchmarks is fundamental to driving research and innovation in a field. SCL is no exception. We briefly discuss existing datasets that can be quickly repurposed for experiments in SCL. In particular, all the datasets listed below can be used within an SCL stream, provided that the evaluation protocol used to optimize the learning models accounts for both quick adaptation and mitigation of forgetting (e.g., by using prequential evaluation and the BWT metric). In general, any dataset including a temporal dimension and shifts in the data distribution is a suitable candidate for an SCL stream.

CLEAR \cite{lin2022a} targets gradual drifts in the temporal evolution of visual concepts. CLEAR can be used for semi-supervised learning, as it includes labelled and unlabeled data adapted from the YFCC100M dataset \cite{thomee2016}. Interestingly, SCL data streams from CLEAR can easily include real drifts. In fact, a given concept changes shapes and characteristics over time, thus requiring updating the learned representations. Together with the introduction of new concepts over time, CLEAR can be readily used to create SCL streams mixing real and virtual drifts. 

Wild-Time~\cite{cit:wild_time} focuses on temporal distribution shifts. Wild-Time provides time stamps for each example, enabling models to build on previously encountered patterns for quick adaptation. The benchmark includes five different real-world datasets on a set of different domains, such as drug discovery, patient prognosis, and news classification. \\
Similarly, CLAD \cite{cit:clad} and KITTI \cite{cit:kitti} capture temporal drifts in vision-based tasks (e.g., object detection). In particular, KITTI includes 6 hours of real-world traffic recorded through various sensors.

CLOC \cite{cai2021} is a dataset of geolocalized, time-stamped images from various locations around the world. The original paper uses CLOC in an OCL setting, therefore being already compatible with SCL approaches.

EGO4D \cite{grauman2024} is a large-scale dataset of daily-life activities from egocentric videos of several users. It is therefore a suitable candidate for personalization tasks where the temporal aspect is key.

Natural language processing is another crucial test-bed for SCL. Like EGO4D, Firehose \cite{hu2023} allows for benchmarking personalization approaches, as it provides tweets from multiple users over time.

TemporalWiki \cite{jang2022} is a dataset of time-stamped Wikipedia and Wikidata pages, allowing for simulation of a data stream where knowledge is continuously updated as some information becomes outdated while new information is included.

Recently, \cite{iovine2025} introduced a benchmark for land use classification from satellite images specifically designed with SCL in mind. Interestingly, predictive models deployed directly on satellites do not have access to large computational resources, making this use case fit for several SCL applications.

\paragraph{Streaming/Continual Reinforcement Learning}
When using the tools developed for CL in CRL, several conceptual problems immediately arise. First of all,
CL techniques are often intrinsically tied to the notion of \textit{tasks}, that is, similarly to the definition of concepts, periods of stationary data statistics whose onsets are known. This idea, when transferred to RL, would imply periods of stationary \textit{environment} statistics. Thus, drift would not occur during tasks, but only at task onsets.
Such a notion is already quite artificial in RL, especially the a priori knowledge about task onsets, and totally incompatible with streaming learning. Task onsets, if they were abrupt, could be detected by dedicated methods, especially for real drifts, where the well-known TD error (the difference between the expected and obtained reward for an action that was executed) should increase. Gradual drift could be detected as well, but this would invalidate the entire concept of a task itself.

The notion of a task is often very useful. For example, replay methods in CL \cite{cossu2025a} partition the memory buffer into fixed-size slots. Each slot gathers examples from the same class (for supervised classification problems) or from the same task (if a task identifier is available). Similarly, replay methods in CRL can exploit information about task labels to partition the replay buffer. However, replay can sometimes be deleterious. In particular, when dealing with an SCL stream where real drifts occur, performing replay would actually preserve the information that the real drifts has now made obsolete.\\

In the absence of a clear separation between tasks in the stream, replay can still work considering boundary-free approaches, like the works in \cite{aljundi2019online, krawczyk24a,krawczyk24b,gepperth2024d}.

\section{A motivating example for SCL}\label{sec:example}
In this Section, we present a simple example to motivate the need for SCL and to empirically prove that SML and CL struggle to find a joint solution to the challenges of quickly adapting to concept drifts while avoiding forgetting. The data and code are available here for \emph{reproducibility}.\footnote{ \url{https://github.com/federicogiannini13/scl-practical-guide}}

\subsection{Experimental setting}
We compare SML and OCL methodologies in two different scenarios by leveraging the MNIST dataset to build our data streams. \\

\begin{figure}[t]
    \centering
    \includegraphics[]{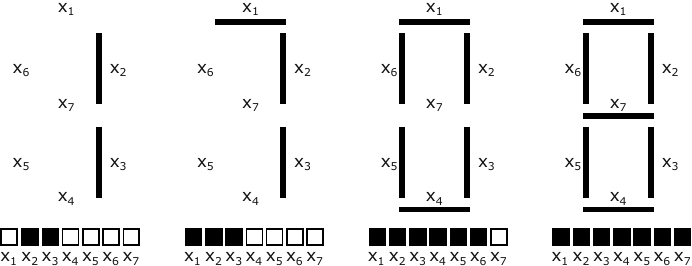}
    \caption{
    Illustration of digits \texttt{1}, \texttt{7}, \texttt{0}, and \texttt{8} using a seven-segment display representation. Each digit is encoded as a binary vector over seven dimensions $(x_1, \dots, x_7)$, where each $x_i$ corresponds to one segment (on = 1, off = 0). Although some digits share subsets of active segments (e.g., \texttt{1} and \texttt{7}), their full encodings place them in distinct regions of the input space. This simplified scenario reflects the nature of virtual drifts in our experiments, where new digits introduce non-overlapping patterns in the input distribution. While MNIST uses grayscale pixel values in a high-dimensional space, the same principle holds: each digit activates a different subset of features, occupying a unique region in the space.
    }
    \label{fig:digits}
\end{figure}

The first scenario (\textbf{virtual drift scenario}) involves only virtual abrupt drifts. Each experience contains images from two digits (one odd digit and one even digit per experience). The learning model is required to classify each digit as odd or even (binary classification). After each experience, a new couple of digits is introduced. This drift alters the data distribution $P(X_t)$.\\
To better understand the nature of virtual drifts, consider a simplified representation of each digit using a 7-dimensional binary input space, similar to the structure of a seven-segment digital display (\cref{fig:digits}). Each dimension corresponds to one segment that can be either on (1) or off (0). For example, the digit \texttt{1} activates only two segments, whereas \texttt{7} adds the top horizontal segment $x_7$. The digit \texttt{0} activates all segments except the central one ($x_7$), while \texttt{8} turns on all seven. In this representation, each digit occupies a distinct region of the input space. Although some digits share identical values on a subset of features. For instance, even if \texttt{1} and \texttt{7} differ only in $x_1$, they still occupy different regions of the input space. In other words, when projected onto the subset of the features defined by $x_2, \dots, x_7$, their representations are indistinguishable, but they lie on different hyperplanes along $x_1$ (e.g., $x_1 = 0$ for \texttt{1}, $x_1 = 1$ for \texttt{7}). Even when digits are similar or lie close in the input space, they are still defined by different combinations of active dimensions and thus reside in separate subregions.

This toy example is not meant to faithfully reproduce the structure of real MNIST images. Instead, it provides an intuitive explanation of how virtual drifts operate. New experiences introduce inputs that occupy previously unseen regions of the input space, while the task definition (odd vs. even classification) remains unchanged. Thus, the labels assigned to the previously observed digits do not change over time. Each drift simply adds new types of instances that extend the problem without contradicting what has been observed so far. The model must learn to extend the previously learned classification rules to previously unexplored regions while retaining the knowledge it has acquired. This example simplifies the much more complex pixel-based input space of MNIST, which operates in grayscale and high dimensionality. In the actual MNIST setting, digits are represented in a much higher-dimensional grayscale pixel space, and although the geometry is far more complex, images belonging to the same digit class still tend to cluster in distinct regions. The simplified 7-segment example is used purely for conceptual illustration and is not used in the experiments.

The second scenario (\textbf{real drift scenario}) involves real abrupt drifts. It consists of five distinct binary classification tasks: (i) even vs. odd, (ii) $>4$ vs. $\leq 4$, (iii) multiple of three vs. not multiple of three, (iv) prime number (including one) vs. non-prime number, (v) $\in[2,5]$ vs. $\notin[2,5]$.\\
Each experience uses the same input distribution, as all examples from all digits are present. However, each experience asks the model to solve a separate task, thus defining a different classification problem. In this case, the distribution $P(X_t)$ remains fixed, while $P(y_t|X_t)$ changes, since the same digit may be assigned different labels depending on the binary task. Additionally, some tasks may share class labels for a subset of the digits.

Our \textbf{research hypotheses} are as follows. \textbf{H1} states that mitigating forgetting slows down the adaptation to new concepts. \textbf{H2} states that SML models are not stable and forget the previous concepts after both real and virtual drifts. \textbf{H3} states that CL strategies struggle to avoid forgetting after real drifts. \textbf{H4} states that CL strategies struggle to efficiently learn the new concept after real drifts. 

To verify these hypotheses, aligning with our SCL perspective, we apply both prequential and CL evaluations. Since some classification problems of the real drift scenarios are imbalanced, we consider Cohen's Kappa Score~\cite{cit:kappa}. Cohen's Kappa is a metric that measures the agreement between predicted and true labels while correcting for the agreement that could occur by chance. Unlike plain accuracy, it provides a more reliable indication of model performance, especially in imbalanced settings. A value of 1 indicates perfect agreement, meaning every prediction is correct; a value of 0 suggests that the model's predictions are no better than random guessing; negative values imply systematic disagreement. In general, values between 0.6 and 0.8 reflect substantial agreement, and those above 0.8 are interpreted as near-perfect, indicating that the model is consistently making accurate predictions beyond chance.

As explained in Section~\ref{sec:sml}, the prequential evaluation updates the score incrementally whenever a new digit $d_t=(X_t,y_t)$ is generated. $X_t$ is the feature vector representing the image, and $y_t$ is the real binary label. The model receives the feature vector $X_t$ and predicts the associated label $\hat{y}_t$. Then, the score is updated considering $y_t$ and $\hat{y}_t$. The prequential evaluation returns an updated score after each data point using a \emph{rolling window} of 1k data points. Thus, at timestamp $t$, it considers the data points from $d_{t-999}$ to $d_t$. After a concept drift, it resets the window to focus on the adaptation to the new concept, avoiding considering the previous concept's last data points. For the first 1k data points following a drift, it only considers the predictions of the data points from the drift onwards. It measures how the model is performing on the recent data points.

During prequential evaluation, a checkpoint of each model is stored after each drift, to be used for the CL evaluation. When the prequential evaluation ends, the CL evaluation computes the average Cohen's Kappa ($K_{avg}$) and backward transfer ($BWT$) metrics using a test set for each concept, which is held out during training (see Section~\ref{sec:cl}). For both metrics, we consider the complete evolution over the stream by averaging across all drifts. Formally, after the concept $i$ ends, the two metrics are computed as in Equations~\ref{eq:k_avg} and \ref{eq:bwt}, where $K_{i,j}$ represents the Cohen's Kappa score achieved by the checkpoint stored after the end of concept $i$ and tested on the test set of concept $j$.

\begin{equation}
    \label{eq:k_avg}
    K_{avg} = \frac{\sum_{i=1}^{N} \sum_{j=1}^{i} K_{i,j}}{\frac{N \cdot (N+1)}{2}},
\end{equation}

\begin{equation}
    \label{eq:bwt}
    BWT = \frac{\sum_{i=2}^{N} \sum_{j=1}^{i-1} (K_{i,j} - K_{j,j})}{\frac{N \cdot (N-1)}{2}} .
\end{equation}

To investigate the effects of CL in streaming scenarios, we evaluate a set of representative replay and regularization strategies. As highlighted in Section~\ref{sec:scl}, these methods are widely used but may struggle with real concept drifts, where conflicting information from past and current tasks can hinder convergence. We consider the following SML models and CL strategies.

\begin{itemize}
    \item ARF~\cite{cit:arf} (see Section~\ref{sec:sml}) is one of the most well-known models in SML. It consists of a streaming version of Random Forest and includes internal drift detectors to autonomously adapt to changes in the data stream, without requiring external drift signals.
    \item Experience Replay (ER) \cite{hayes2021} is a widely used strategy in CL that mitigates forgetting by storing a small subset of past examples in a memory buffer and periodically revisiting them during training, allowing the model to reinforce previous knowledge while learning new tasks.
    \item AGEM~\cite{cit:agem} prevents forgetting by projecting the current gradient update onto the closest direction that does not increase the loss on a small memory of past examples. Instead of retraining directly on stored data, it uses these samples to constrain updates, ensuring that new learning does not interfere with previously acquired knowledge.
    \item Naive classifier: it is the base learning model without any CL strategy. The model is fine-tuned over the data stream continuously.
\end{itemize} 

The methodologies' choice is directly tied to the nature of real drifts and to the hypotheses formulated in this work. Replay and regularization CL strategies (represented by ER and AGEM) are included because they jointly optimize past and current tasks. While this approach is effective in virtual drift scenarios, it is fundamentally misaligned with real drifts, where past and current concepts may be incompatible. Under these conditions, the gradient signals introduced by replay or regularization can become contradictory, preventing convergence and slowing down the adaptation process. For this reason, ER and AGEM provide ideal representatives to test whether these mechanisms fail under real drift, as predicted by H1, H3, and H4.

The Naive classifier is included as a fully plastic model without any strategy to avoid forgetting. When compared to CL strategies, it represents a lower-bound baseline for stability evaluation and an upper-bound baseline for plasticity evaluation. In contrast, ARF is selected as a representative of SML approaches due to its well-established ensemble mechanism and robustness in streaming settings.

CL learning strategies are trained using the OCL approach with mini-batches containing ten images. The concept boundaries are given to the model. During each step of the prequential evaluation, ARF performs inference on a single data point and is trained on the same data point. OCL methods follow the same protocol, but with minibatches of 10 examples at a time. No method performs multiple passes on the same data points. To perform a fair comparison between decision tree-based and deep learning-based models, images are represented using a common feature space of 30 components given by a UMAP model~\cite{cit:umap}. The UMAP model is trained offline on the whole MNIST dataset.

OCL strategies are implemented using the Avalanche library\cite{carta2023}\footnote{\url{https://avalanche.continualai.org/}} with a base learner consisting of a neural network with one hidden layer and size 512. All the models use a Stochastic Gradient Descent optimizer with learning rate $0.001$ and momentum $0.9$.\footnote{\url{https://docs.pytorch.org/docs/stable/generated/torch.optim.SGD.html}} ARF is implemented using the River library with default parameters\footnote{\url{https://riverml.xyz/0.11.1/api/ensemble/AdaptiveRandomForestClassifier/}}. ER uses a replay buffer of ten data points for each mini-batch, sampled from a memory of 500 images.

\subsection{Results}

\begin{figure}[t]
    \centering
    \begin{subfigure}{\textwidth}
        \centering
        \includegraphics{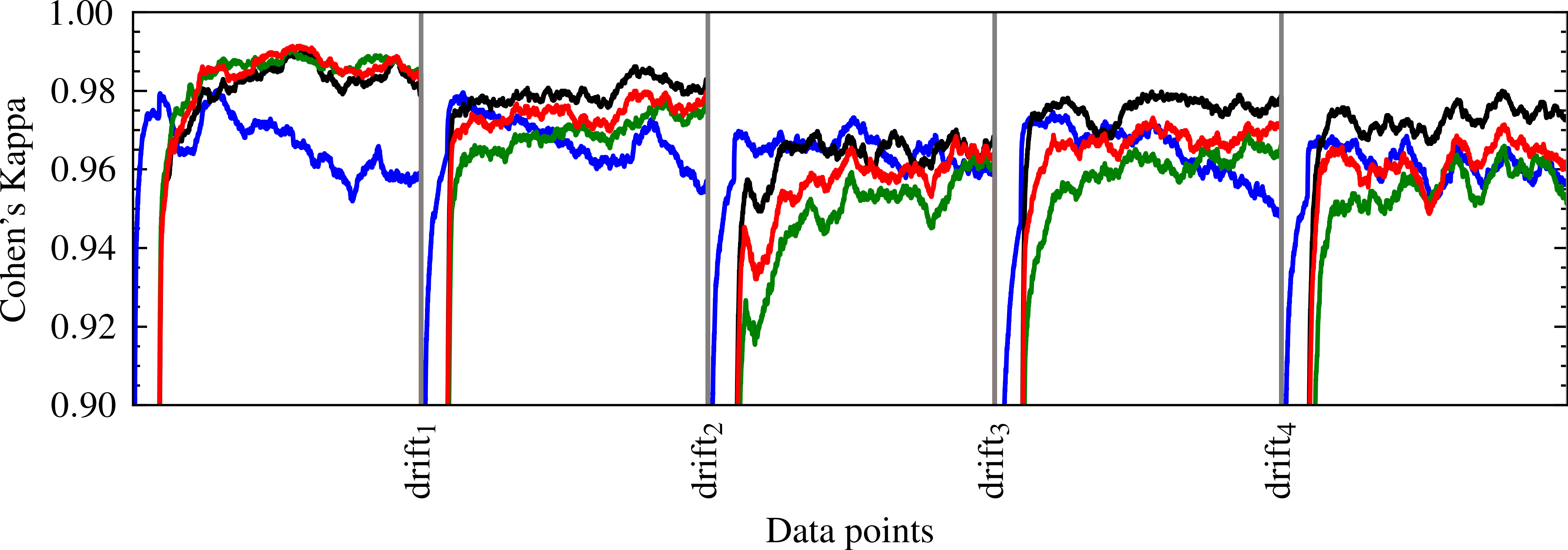}
        \caption{Virtual drift scenario}
        \label{fig:prequential_ev_virtual}
    \end{subfigure}

    \begin{subfigure}{\textwidth}
        \centering
        \includegraphics{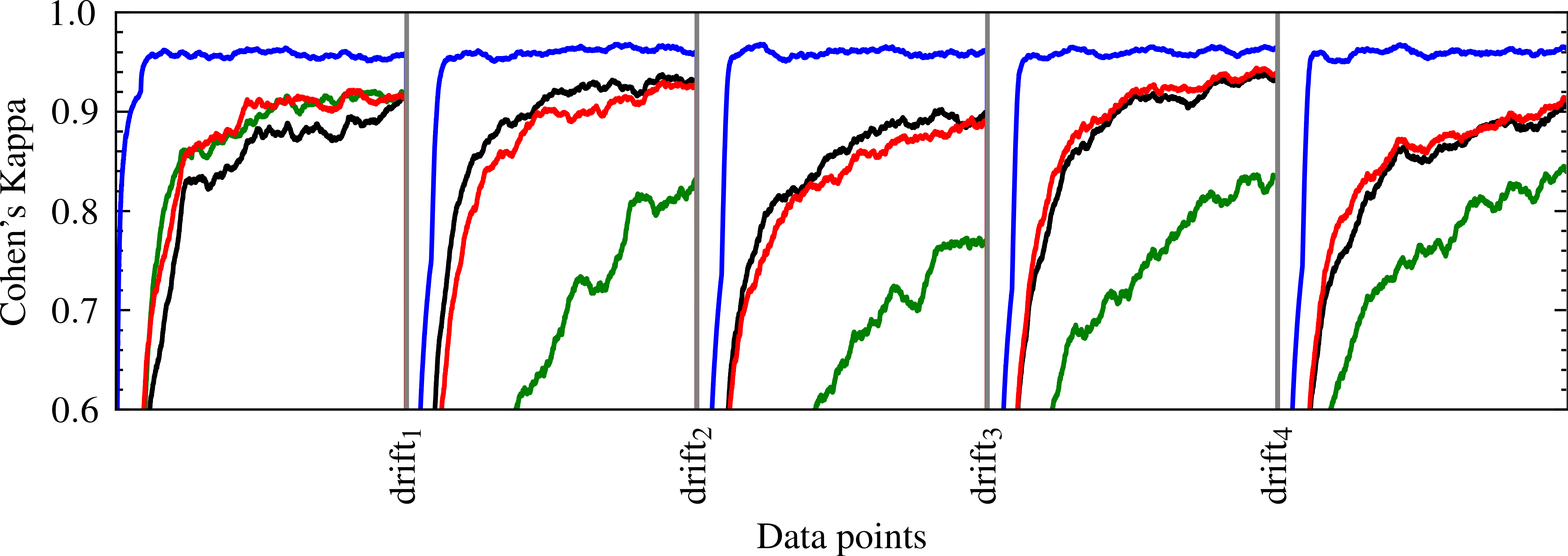}
        \caption{Real drift scenario}
        \label{fig:prequential_ev_real}
    \end{subfigure}

    \begin{subfigure}{\textwidth}
        \centering
        \includegraphics{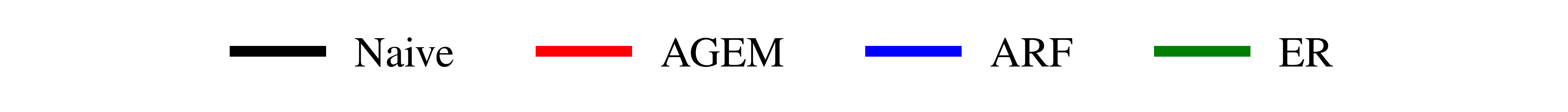}
        \phantomcaption{}
        \label{fig:performance_averaged_aq}
    \end{subfigure}
    \caption{Prequential evaluation results averaged over the ten executions (considering Cohen's Kappa score). In the virtual drift scenario (\ref{fig:cl_ev_virtual}), the Naive classifier performs well. ER and AGEM obtain slightly lower performance, since they balance between plasticity and stability. In the real drift scenarios (\ref{fig:cl_ev_real}), the Naive classifier and AGEM  perform equally, since the AGEM strategy has no impact. ER has strong difficulties to learn the new concept since it tries to mitigate forgetting. ARF is plastic in both scenarios.}
    \label{fig:prequential_ev}
\end{figure}
For the virtual drift scenario, we average results across ten runs with different orders and pairs of digits. For the real drift scenario, we follow the same protocol by changing the order of the classification tasks.

\begin{figure}[t]
    \centering

    \begin{subfigure}{0.48\textwidth}
        \centering
        \includegraphics{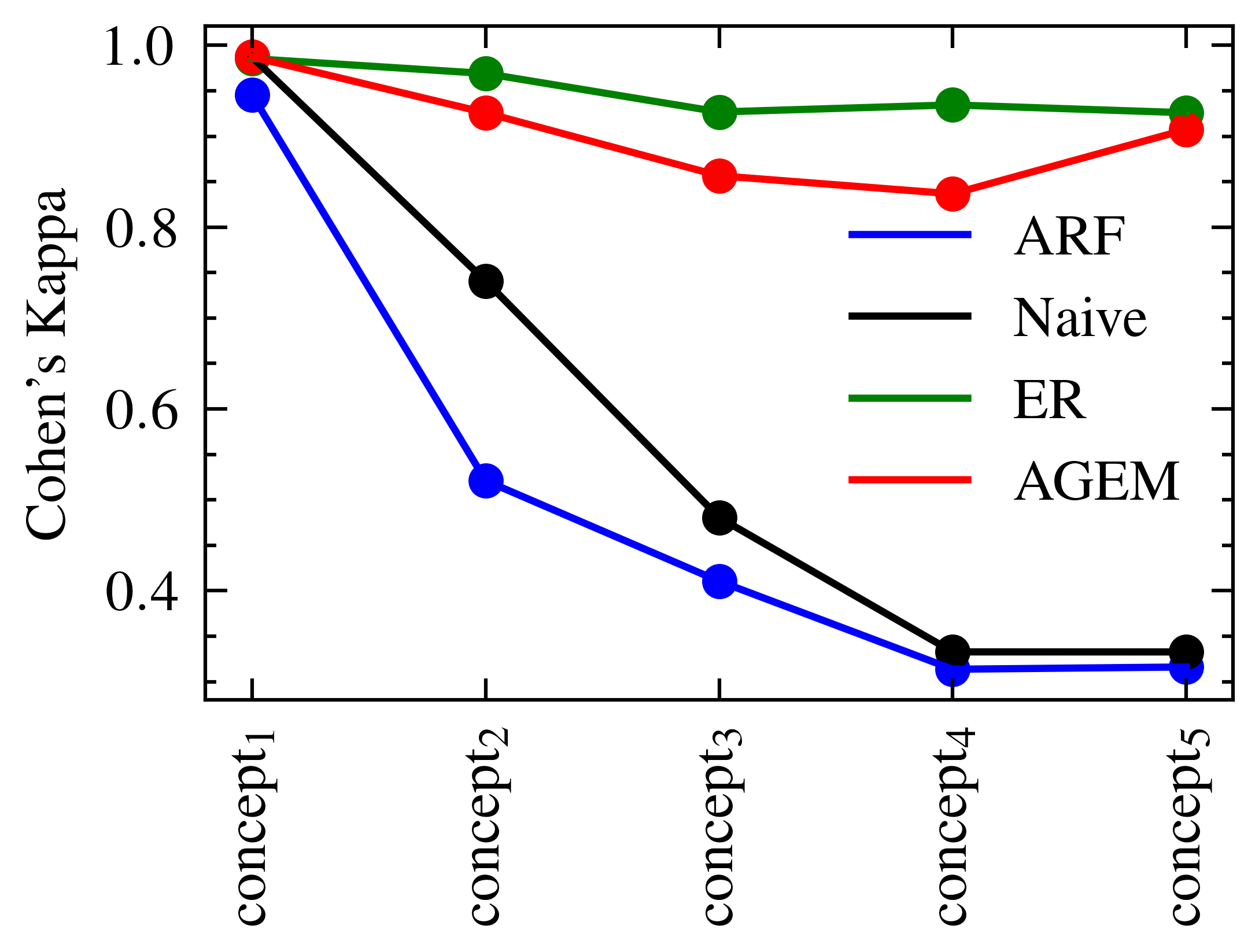}
        \caption{Virtual drift scenario}
        \label{fig:cl_ev_virtual}
    \end{subfigure}%
    \hfill
    \begin{subfigure}{0.48\textwidth}
        \centering
        \includegraphics{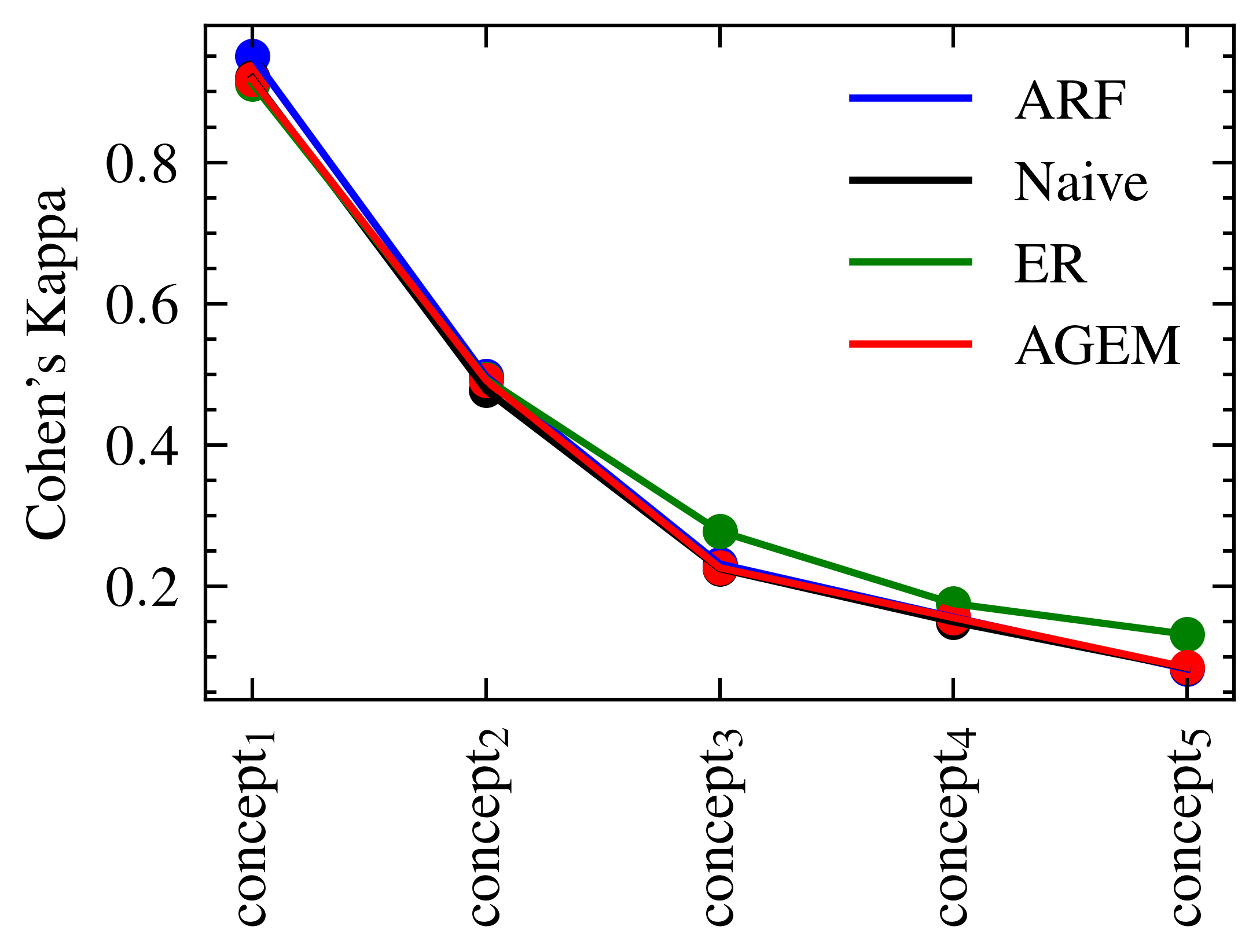}
        \caption{Real drift scenario}
        \label{fig:cl_ev_real}
    \end{subfigure}
    \caption{CL learning evaluation. After each drift, the models are tested over the current and all the previous test sets, and the average Cohen's Kappa score is computed. Finally, results are averaged over the ten executions. On the virtual drift scenario (\ref{fig:cl_ev_virtual}), ER and AGEM are stable, while ARF and the Naive classifier catastrophically forget previous concepts. On the virtual drift scenario (\ref{fig:cl_ev_real}), all the models cannot avoid forgetting. This is due to the fact that the concepts are in contradiction, and the model cannot solve both of them jointly.}
    \label{fig:cl_ev}
\end{figure}

Results are shown in \cref{fig:prequential_ev} (prequential evaluation) and \cref{fig:cl_ev} (CL evaluation). We recall that the main difference lies in the fact that the CL evaluation tests the models after each drift on the test sets of the concepts encountered so far. Table~\ref{table:cl_ev} shows the results for $K_{avg}$ and $BWT$ metrics.\\

When analyzing the virtual drift scenario, the prequential evaluation (\cref{fig:prequential_ev_virtual}) reveals that the decision tree model performs worse than the neural network. ARF, in fact, generally achieves lower prequential accuracy than the Naive classifier. ARF can adapt more quickly to the very first data points following a drift, whereas the Naive classifier performs very poor. However, it is important to note that the Naive classifier is trained incrementally on mini-batches, which introduces a delay in adapting to new data. Both ER and AGEM are outperformed by the Naive classifier, indicating that applying specific CL strategies can slightly hinder the learning of the current concept. This outcome is further clarified by the CL evaluation (\cref{fig:cl_ev_virtual} and Table~\ref{table:cl_ev}). ER exhibits the best performance in terms of mitigating forgetting, highlighting its stability. However, this stability comes at the expense of reduced plasticity, in line with the well-known stability-plasticity trade-off \cite{carpenter1986}. In contrast, Naive and ARF show strong online performance but suffer from catastrophic forgetting of previous concepts, since they do not apply specific strategies to mitigate it. AGEM, while slightly less stable than ER, achieves better online performance and approaches the Naive classifier.\\

Results are significantly different in the real drift scenario. As shown in the prequential evaluation (\cref{fig:prequential_ev_real}), decision trees outperform neural networks. AGEM performs on par with the Naive classifier, while ER struggles to learn the current concept effectively. This outcome is further explained by the CL evaluation results (\cref{fig:cl_ev_real} and Table~\ref{table:cl_ev}). AGEM and the Naive classifier achieve similarly poor scores, indicating that AGEM fails to provide any meaningful improvement and behaves essentially like a Naive classifier. ARF once again catastrophically forgets all previously learned concepts. ER performs slightly better in terms of retaining past knowledge, but still fails to prevent forgetting effectively. The underlying issue lies in ER's training approach: it mixes the current mini-batch with replayed examples from past concepts. However, in the presence of real drifts, these past examples are associated with previous decision boundaries that contradict the new target concept. As a result, the model receives contradictory training signals, which not only prevent it from preserving old knowledge but also hinder its ability to learn the new concept. Consequently, training fails to converge, and ER is unable to adapt properly in this setting.\\

\begin{table}[t]
\centering
\begin{tabular}{r|ll|ll|}
\toprule
 & \multicolumn{2}{c|}{\textbf{Virtual drift scenario}} & \multicolumn{2}{c|}{\textbf{Real drift scenario}} \\
 & \multicolumn{1}{c}{$\mathbf{K_{avg}}$} & \multicolumn{1}{c|}{$\mathbf{BWT}$} & \multicolumn{1}{c}{$\mathbf{K_{avg}}$} & \multicolumn{1}{c|}{$\mathbf{BWT}$} \\ \midrule
\textbf{AGEM} & 0.89 ± 0.06 & -0.13 ± 0.10 & 0.24 ± 0.01 & -0.99 ± 0.05 \\ 
\textbf{ARF} & 0.40 ± 0.09 & -0.83 ± 0.14 & 0.24 ± 0.00 & -1.06 ± 0.00 \\ 
\textbf{ER} & 0.94 ± 0.02 & -0.04 ± 0.02 & 0.27 ± 0.02 & -0.85 ± 0.08 \\ 
\textbf{Naive} & 0.46 ± 0.09 & -0.77 ± 0.14 & 0.24 ± 0.01 & -1.00 ± 0.04 \\ \bottomrule
\end{tabular}
\caption{Mean and standard deviation of the CL metrics over the ten executions using Cohen's Kappa score.}
\label{table:cl_ev}
\end{table}

To summarize, all the research hypotheses are verified. \textbf{H1:} the Naive classifier is more plastic than CL strategies. \textbf{H2:} ARF exhibits a poor CL performance in both scenarios. \textbf{H3:} in the real drift scenario, AGEM and ER cannot avoid forgetting since previous concepts are in contradiction with the current ones, and the model cannot solve all of them jointly. \textbf{H4:} in the real drift scenario, AGEM is not effective and performs on par with the Naive classifier. Similarly, ER struggles with learning the current concept.

\section{Conclusion and Future Directions}\label{sec:conclusion}
We discussed the Streaming Continual Learning scenario, where a model is required to quickly adapt to new information while retaining previous knowledge. We motivated the need for SCL with an empirical evaluation of existing SML and CL strategies, briefly showing how the two objectives are currently not met by any of the approaches alone (Section \ref{sec:example}). While CL focuses on knowledge consolidation to avoid forgetting, SML focuses on rapid adaptation. This also results in different evaluation protocols, hiding the opportunities for a positive exchange of ideas and solutions. \\
We do not believe that SCL will replace SML or CL. Rather, we strive for an integration of the two approaches. Throughout this paper, we claimed that such opportunities are already available, and that they are currently being neglected. \\
While SCL ideas are gaining traction along several directions (e.g., online continual learning, SML with deep representation learning), a unified approach is still missing. Bringing together the SML and CL communities will speed up the process. \\

We foresee potential applications of SCL in areas where i) temporal correlations play a role and ii) virtual and real drift can occur at any time.

The presence of \emph{temporal dependencies} influences many real-world scenarios, including the Internet of Things, robotics, object tracking, video surveillance, and satellite imagery analysis~\cite{cossu2025a}. Despite its crucial relevance~\cite{cit:tenet}, this problem is usually ignored by both research areas. As highlighted by M5 challenge~\cite{cit:makridakis_m5}, combining statistical and deep learning approaches can be effective in modelling temporal patterns. Moving towards an SCL perspective, integrating SML and CL methodologies could provide a viable solution. Since Recurrent Neural Networks models are widely applied in this case, they can also be used online in an SML way to learn continuously and handle temporal dependencies. Furthermore, architectural CL strategies may be applied on top of them to avoid forgetting and exploit transfer learning. Continuous Progressive Neural Networks~\cite{cit:cpnn} represent one of the first pioneering solutions that apply this rationale. 

One example satisfying both temporal dependence and the presence of real and virtual drifts is the case of \emph{personalization}. When a predictive model needs to be tailored to each user's needs, it is important to consider how the user's preferences evolve over time, what is relevant, and what is instead becoming outdated.

Equipped with the intuitions and the results from Section~\ref{sec:example}, SCL approaches would enable rapid adaptation to novel needs (virtual drifts), while at the same time preventing forgetting of old preferences. Selective forgetting (e.g., via model editing or similar techniques) would still be required in the presence of real drift, where old preferences become outdated. As we have seen, carefully balancing these two dimensions is still a challenge for existing approaches, and one of the most important and pressing motivations behind the study of SCL scenarios.

\section*{Acknowledgments}
This work has been partially funded by the EU-EIC EMERGE project (Grant No. 101070918) and by the HEU CoEvolution project (Grant No. 101168560).

\bibliographystyle{elsarticle-num} 
\bibliography{bib}

\end{document}